\documentclass[sigconf,authorversion]{acmart}
\acmSubmissionID{524}

\usepackage{amsmath}
\usepackage{bbm}
\usepackage{booktabs} 
\usepackage[commandnameprefix=always,authormarkup=none,final]{changes}
\usepackage{enumitem}
\usepackage{epsfig}
\usepackage{graphicx}
\usepackage{makecell}
\usepackage{multirow}
\usepackage{subcaption}
\usepackage{xcolor}

\usepackage[ruled]{algorithm2e} 

\SetAlFnt{\small}
\SetAlCapFnt{\small}
\SetAlCapNameFnt{\small}
\SetAlCapHSkip{0pt}

\acmJournal{TOG}

\copyrightyear{2023}
\acmYear{2023}
\setcopyright{acmlicensed}\acmConference[SA Conference Papers '23]{SIGGRAPH Asia 2023 Conference Papers}{December 12--15, 2023}{Sydney, NSW, Australia}
\acmBooktitle{SIGGRAPH Asia 2023 Conference Papers (SA Conference Papers '23), December 12--15, 2023, Sydney, NSW, Australia}
\acmPrice{15.00}
\acmDOI{10.1145/3610548.3618188}
\acmISBN{979-8-4007-0315-7/23/12}

\usepackage{amsfonts}
\usepackage{textcomp}

\usepackage[capitalize]{cleveref}
\crefname{section}{Sec.}{Secs.}
\Crefname{section}{Section}{Sections}
\Crefname{table}{Table}{Tables}
\crefname{table}{Tab.}{Tabs.}

\usepackage{scalerel}
\newcommand\Tau{\scalerel*{\tau}{T}}

\definechangesauthor[name=new additions, color=teal]{new}
\definechangesauthor[name=in place modifications, color=blue]{modified}
\definechangesauthor[name=deletions, color=red]{deleted}

\newcommand{\losscolor}{{\mathcal{L}_{\text{color}}}}
\newcommand{\lossap}{{\mathcal{L}_{\text{ap}}}}
\newcommand{\lossav}{{\mathcal{L}_{\text{av}}}}
\newcommand{\losscfc}{{\mathcal{L}_{\text{cfc}}}}
\newcommand{\losssd}{{\mathcal{L}_{\text{sd}}}}

\newcommand{\pixelq}{\mathbf{q}}
\newcommand{\pointp}{\mathbf{p}}
\newcommand{\viewv}{\mathbf{v}}
\newcommand{\colorc}{\mathbf{c}}
\newcommand{\featureh}{\mathbf{h}}
\newcommand{\mlpf}{\mathcal{F}}
\newcommand{\indicatorfunc}{\mathbbm{1}}

\begin{document}
    \title{SimpleNeRF: Regularizing Sparse Input Neural Radiance Fields with Simpler Solutions}

\author{Nagabhushan Somraj}
\orcid{0000-0002-2266-759X}
\affiliation{%
  \institution{Indian Institute of Science}
  \city{Bengaluru}
  \state{Karnataka}
  \postcode{560012}
  \country{India}}
\email{nagabhushans@iisc.ac.in}
\author{Adithyan Karanayil}
\affiliation{%
  \institution{Indian Institute of Science}
  \city{Bengaluru}
  \country{India}
}
\email{adithyanv@iisc.ac.in}
\author{Rajiv Soundararajan}
\affiliation{%
 \institution{Indian Institute of Science}
 \city{Bengaluru}
 \country{India}}
\email{rajivs@iisc.ac.in}


    \begin{abstract}
        Neural Radiance Fields (NeRF) show impressive performance for the photo-realistic free-view rendering of scenes.
        However, NeRFs require dense sampling of images in the given scene, and their performance degrades significantly when only a sparse set of views are available.
        Researchers have found that supervising the depth estimated by the NeRF helps train it effectively with fewer views.
        The depth supervision is obtained either using classical approaches or neural networks pre-trained on a large dataset.
        While the former may provide only sparse supervision, the latter may suffer from generalization issues.
        As opposed to the earlier approaches, we seek to learn the depth supervision by designing augmented models and training them along with the NeRF.
        We design augmented models that encourage simpler solutions by exploring the role of positional encoding and view-dependent radiance in training the few-shot NeRF. 
        The depth estimated by these simpler models is used to supervise the NeRF depth estimates. 
        Since the augmented models can be inaccurate in certain regions, we design a mechanism to choose only reliable depth estimates for supervision.
        Finally, we add a consistency loss between the coarse and fine multi-layer perceptrons of the NeRF to ensure better utilization of hierarchical sampling.
        We achieve state-of-the-art view-synthesis performance on two popular datasets by employing the above regularizations.
        The source code for our model can be found on our project page: \url{https://nagabhushansn95.github.io/publications/2023/SimpleNeRF.html}
    \end{abstract}

%
%


%
%


    \begin{teaserfigure}
        \centering
        \begin{subfigure}{0.48\linewidth}
            \centering
            \includegraphics[width=\linewidth]{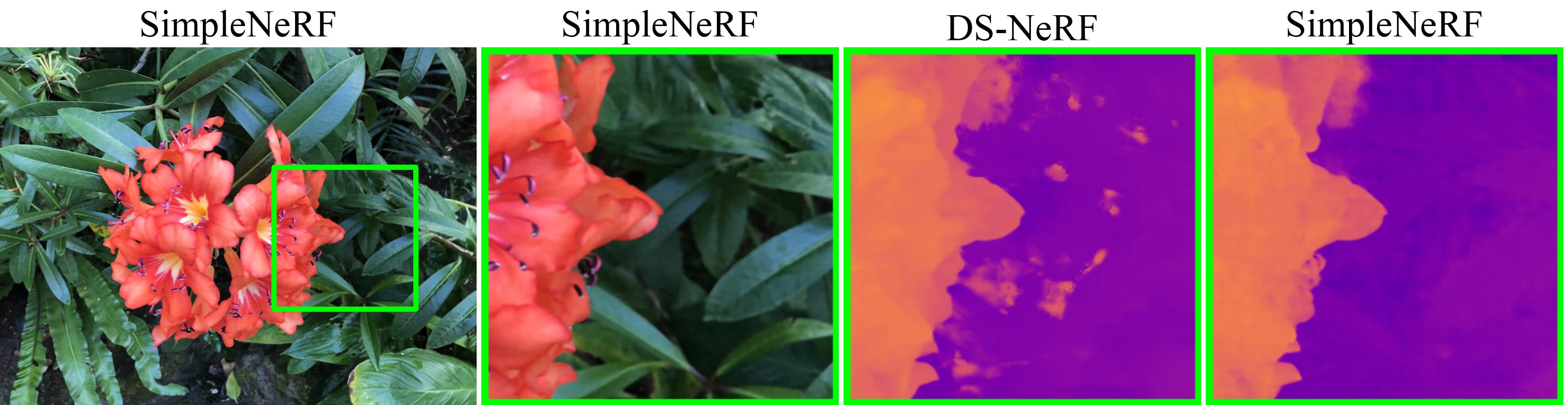}
            \caption{\textbf{Mitigation of floaters:}
            For a frame from the LLFF flower scene, we show the depths predicted by DS-NeRF and SimpleNeRF.
            We show the complete frame for reference and focus on a small region to better observe the floaters.
            }
            \label{fig:teaser-points-augmentation}
        \end{subfigure}
        \hfill
        \begin{subfigure}{0.48\linewidth}
            \centering
            \includegraphics[width=\linewidth]{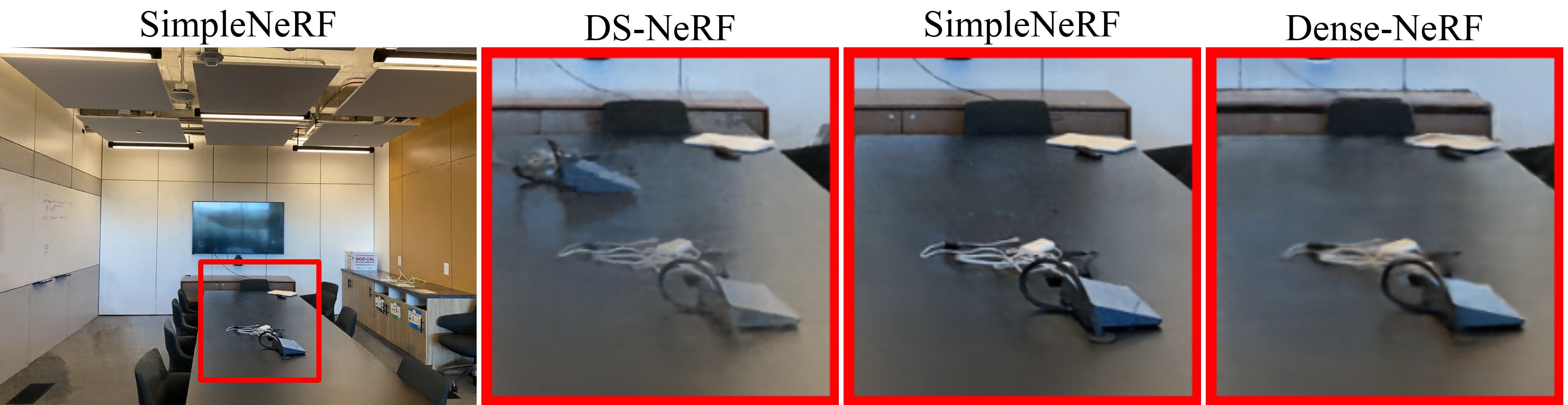}
            \caption{\textbf{Reducing shape-radiance ambiguity:}
            This is an example from the LLFF room scene.
            For reference, we show the prediction of Vanilla NeRF trained with dense input views.
            }
            \label{fig:teaser-views-augmentation}
        \end{subfigure}
        \begin{subfigure}{\linewidth}
            \centering
            \begin{subfigure}{0.48\linewidth}
                \centering
                \includegraphics[width=\linewidth]{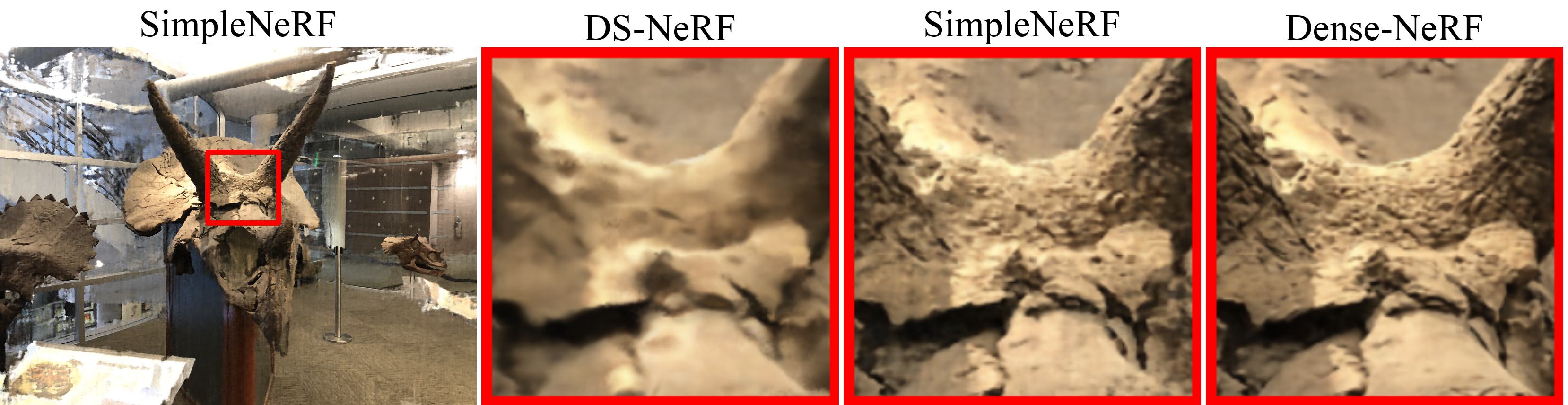}
                \caption{\textbf{Improving Sharpness:}
                The above images correspond to the LLFF horns scene.
                We enlarge a small region of the frame to better observe the improvement in sharpness.
                }
                \label{fig:teaser-hierarchical-sampling2}
            \end{subfigure}
            \hfill
            \begin{subfigure}{0.48\linewidth}
                \centering
                \includegraphics[width=\linewidth]{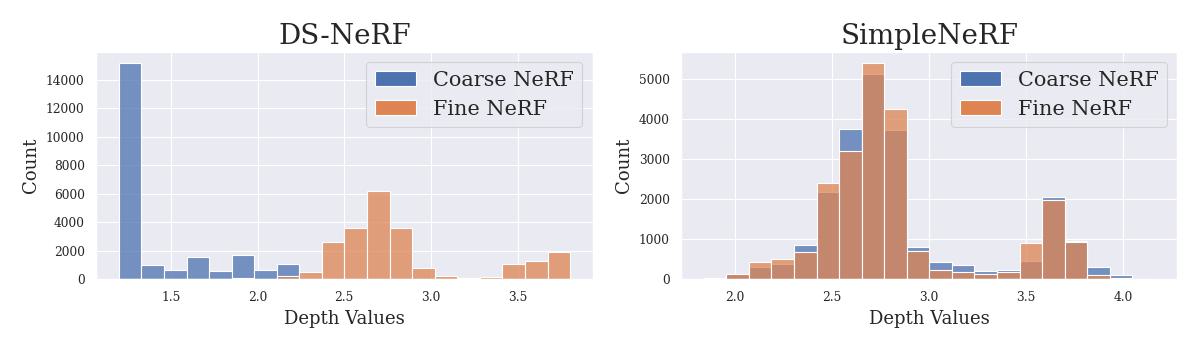}
                \caption{\textbf{Improving Sharpness:}
                Histogram of depth values predicted by the coarse and fine NeRF models for the image patch shown in Fig (c).
                }
                \label{fig:teaser-hierarchical-sampling1}
            \end{subfigure}
        \end{subfigure}
        \caption{We show three shortcomings of DS-NeRF when trained with two input views on the LLFF dataset.
        SimpleNeRF introduces regularizations on DS-NeRF to mitigate these distortions.
        Notice the floaters in DS-NeRF predictions shown by small orange regions in Fig. (a) which are cleaned by SimpleNeRF.
        In Fig. (b), we find DS-NeRF suffers from shape-radiance ambiguity and ends up changing the color of the table in different viewpoints to match the observed images.
        DS-NeRF blends the two input images based on the viewpoint instead of learning correct geometry.
        This leads to ghosting artifacts in novel viewpoints, which is removed by our model.
        In Fig (c), we observe SimpleNeRF producing sharper reconstructions than DS-NeRF.
        Fig (d) shows a possible reason, where the coarse and fine NeRF models in DS-NeRF converge to different depth estimates.
        This leads to ineffective hierarchical sampling resulting in blurry predictions.
        We find that SimpleNeRF mitigates this by predicting consistent depth estimates.
        }
        \label{fig:teaser}
    \end{teaserfigure}

    \maketitle

    \section{Introduction}\label{sec:introduction}
    Neural Radiance Fields (NeRFs)~\cite{mildenhall2020nerf} show unprecedented levels of performance in synthesizing novel views of a scene by learning a volumetric representation implicitly within the weights of multi-layer perceptrons (MLP).
    Although NeRFs are very promising for view synthesis, there is a need to improve their design in a wide array of scenarios.
    For example, NeRFs have been enhanced to learn on unbounded scenes~\cite{barron2022mipnerf360}, handle scenes with highly specular objects~\cite{verbin2022refnerf}, and deal with noisy camera poses~\cite{bian2023nopenerf}.
    Yet NeRFs require tens to hundreds of images per scene to learn the scene geometry accurately, and their quality deteriorates significantly when only a few training images are available~\cite{jain2021dietnerf}.
    In this work, we focus on training the NeRF with a sparse set of input images and aim to design novel regularizations for effective training.

    Prior work on sparse input NeRFs can be classified into generalized models and regularization based models.
    Generalized models~\cite{johari2022geonerf} use convolutional neural networks to obtain a latent representation of the scene and use it to condition the NeRF\@.
    However, these models require a large dataset for pre-training and may suffer from generalization issues when used to render a novel scene~\cite{niemeyer2022regnerf}.
    The other thread of work on sparse input NeRFs follows the original NeRF paradigm of training scene-specific NeRFs, and designs novel regularizations to assist NeRFs in converging to a better scene geometry~\cite{zhang2021ners}.
    One popular approach among such models is to supervise the depth estimated by the NeRF.
    RegNeRF~\cite{niemeyer2022regnerf}, DS-NeRF~\cite{deng2022dsnerf} and ViP-NeRF~\cite{somraj2023vipnerf} use simple priors such as depth smoothness, sparse depth or relative depth respectively obtained through classical approaches.
    On the other hand, DDP-NeRF~\cite{roessle2022ddpnerf} and SCADE~\cite{uy2023scade} pre-train convolutional neural networks (CNN) on a large dataset of scenes to learn a prior on the depth or its probability distribution.
    These approaches may also suffer from similar issues as the generalized models.
    This raises the question of whether we can instead learn the depth supervision in-situ without employing any pre-training.
    We aim to regularize NeRFs by learning augmented models for depth supervision in tandem with the NeRF training.

    We follow the popular Occam's razor principle and regularize the NeRF by designing augmented models to choose simpler solutions over complex ones, wherever possible.
    Some of the key components of the NeRF such as positional encoding, view-dependent radiance, and hierarchical sampling provide powerful capabilities to the NeRF and are designed for the case with dense input views.
    Existing implementations of these components may be sub-optimal with fewer input views, perhaps due to the highly under-constrained system, causing several distortions.
    \cref{fig:teaser} shows three common distortions, namely, floaters~\cite{roessle2022ddpnerf}, shape-radiance ambiguity~\cite{zhang2020nerfpp}, and blurred renders for the NeRF model in the few-shot setting.
    We believe that by simplifying some of the capabilities, one can obtain simpler augmented models that may provide better depth supervision for training the NeRF\@.
    

    We simplify the capability of the augmented NeRF models with respect to positional encoding and view-dependent radiance.
    Positional encoding maps two nearby points in 3D space to far-apart points in the encoded space.
    This allows the NeRF to learn sharp depth discontinuities in 3D space via a smooth function in the encoded space.
    However, in an under-constrained system, the NeRF learns many undesirable depth discontinuities that can still explain the observed images.
    This leads to floater artifacts as seen in \cref{fig:teaser-points-augmentation}.
    Restricting the ability of the NeRF to learn these sharp discontinuities encourages it to mitigate the floater artifacts.
    Similarly, the ability of the NeRF to predict view-dependent radiance leads to shape-radiance ambiguity, which we find to be more pronounced in the case of few-shot NeRF\@.
    The NeRF can learn to explain the observed images by modifying the color of 3D points in accordance with the input images as seen in \cref{fig:teaser-views-augmentation}.
    Restricting the NeRF to predict view-independent radiance only encourages the NeRF to explain the observed images using simpler Lambertian surfaces and can help avoid shape-radiance ambiguity.

    We use the depth estimated by the simpler augmented models to supervise the depth estimated by the NeRF model.
    Note that we use the simplified models as augmentations for depth supervision and not as the main NeRF model since na\"{\i}vely reducing the capacity of the NeRF may lead to suboptimal solutions~\cite{jain2021dietnerf}.
    For example, the model that predicts view-independent radiance fails if the scene contains specular objects.
    Further, the simpler augmented models need to be used for supervision only if they explain the observed images accurately.
    We gauge the reliability of the depths by reprojecting pixels using the estimated depths onto a different nearest train view and comparing them with the corresponding images.
    We use DS-NeRF~\cite{deng2022dsnerf} as our baseline and design our regularizations on top of it.
    Our framework can thus be seen as a semi-supervised learning model by considering the sparse depth from a Structure from Motion (SfM) module as providing limited depth labels and the remaining pixels as the unlabeled data.
    Our approach of using augmented models in tandem with the main NeRF model is perhaps closest to the Dual-Student architecture~\cite{ke2019dualstudent} that trains another identical model in tandem with the main model and imposes consistency regularization between the predictions of the two models.
    In contrast, our augmented models have complementary abilities as compared to the main NeRF model.

    Finally, we observe that the coarse and fine NeRFs may converge to different depth estimates when trained with fewer images, as seen in \cref{fig:teaser-hierarchical-sampling1}.
    This essentially renders the hierarchical sampling ineffective.
    The resulting model is similar to the one with undersampled points along the rays.
    We avoid such degenerate cases by imposing a consistency loss between the depths estimated by the coarse and fine MLPs.
    We achieve state-of-the-art view synthesis performance on two popular datasets with the above three regularizations.
    Further, we show that our model learns significantly improved geometry as compared to the prior art.
    We refer to our model as SimpleNeRF\@.

    We list the main contributions of our work in the following.
    \begin{itemize}
        \item We design two augmented NeRF models that are biased towards simpler solutions and use the depth estimates from these models to regularize the NeRF training.
        Through the two augmentations, we mitigate incorrect depth discontinuities and shape-radiance ambiguity.
        \item We design a mechanism to determine whether the depths estimated by the augmented models are accurate and utilize only the accurate estimates to supervise the NeRF.
        \item We improve the effectiveness of hierarchical sampling by introducing a consistency constraint between the coarse and fine NeRFs. 
        This generates sharper frames.
        \item We achieve the state-of-the-art performance of few shot NeRFs on two popular datasets.
    \end{itemize}

    \section{Related Work}\label{sec:related-work}
    Novel view synthesis is a classic problem traditionally solved broadly using image based rendering~\mbox{\cite{chen1993view}} or light fields~\mbox{\cite{gortler1996lumigraph,levoy1996light}}.
    The seminal work by ~\citet{mildenhall2020nerf} started a new pathway in neural view synthesis and led to NeRF based models being employed in a wide variety of applications such as 3D editing of scenes~\cite{yuan2022nerfediting}, gaming~\cite{menapace2023plotting}, extended reality~\cite{deng2022fovnerf} and image reconstruction~\cite{mildenhall2022nerfinthedark,pearl2022nan,ma2022deblurnerf}, among others.
    However, many of the above models require dense sampling of input views for a faithful 3D geometry generation.
    With fewer views, the quality of rendered novel views and the learned 3D geometry degrade significantly introducing severe distortions.
    This can limit the widespread usage of NeRFs in multiple applications and hence addressing this limitation is of considerable interest.

    \subsection{Generalized Sparse Input NeRF}\label{subsec:generalized-sparse-input-nerf}
    Prior work involves various approaches to learning a 3D neural representation with fewer input views.
    One line of approach attempts to train a generalized model on a large dataset of scenes such that the model can utilize the learned prior to generate a 3D scene representation from the few input images~\cite{chen2021mvsnerf,tancik2021metanerf,lee2023extremenerf}.
    Early pieces of work such as PixelNeRF~\cite{yu2021pixelnerf}, GRF~\cite{trevithick2021grf}, and IBRNet~\cite{wang2021ibrnet} obtain convolutional features of the input images and additionally condition the NeRF by projecting the 3D points onto the feature grids.
    MVSNeRF~\cite{chen2021mvsnerf} and SRF~\cite{chibane2021srf} incorporate cross-view knowledge into the features by constructing a cost volume and pair-wise post-processing of the individual frame features respectively.
    GeoNeRF~\cite{johari2022geonerf} further improves the performance by employing a transformer to effectively reason about the occlusions in the scene.
    More recent work such as GARF~\cite{shi2022garf} and MatchNeRF~\cite{chen2023explicit} try to provide explicit knowledge about the scene geometry through depth maps and similarity of the projected features respectively.
    This approach of conditioning the NeRF on learned features is also popular among single image NeRF models~\cite{lin2023vision}, which can be considered an extreme case of sparse input NeRF\@.
    However, the need for pre-training on a large dataset of scenes with multi-view images and issues due to domain shift have motivated researchers to adopt regularization based approaches.

    \subsection{Regularization based Sparse Input NeRF}\label{subsec:regularized-sparse-input-nerf}
    A popular approach among regularization based models is to regularize the depth estimated by the NeRF\@.
    DS-NeRF~\cite{deng2022dsnerf} uses sparse depth provided by a SfM module to supervise the NeRF estimated depth at sparse keypoints.
    RegNeRF~\cite{niemeyer2022regnerf} imposes the depth smoothness constraint on the rendered depth maps.
    ViP-NeRF~\cite{somraj2023vipnerf} instead attempts to regularize the relative depth of objects by obtaining a prior on the visibility of objects.
    Unlike the above, few other works exploit the advances in depth-estimation using deep neural networks.
    DDP-NeRF~\cite{roessle2022ddpnerf} extends DS-NeRF by employing a CNN to convert the sparse depth into dense depth for more supervision.
    SCADE~\cite{uy2023scade} and SparseNeRF~\cite{wang2023sparsenerf} use the depth map output by single image depth models to constrain the absolute and the relative order of pixel depths respectively.
    DiffusioNeRF~\cite{wynn2023diffusionerf} learns the joint distribution of RGBD patches using denoising diffusion models (DDM) and utilizes the gradient of the distribution provided by the DDM to regularize NeRF rendered RGBD patches.
    Different from the above, our work obtains depth supervision by harnessing the power of learning through augmented models, but without the need for pre-training on a large dataset.

    Another line of regularization based approaches hallucinate new viewpoints and regularize the NeRF on different aspects such as semantic consistency~\cite{jain2021dietnerf}, depth smoothness~\cite{niemeyer2022regnerf}, sparsity of mass~\cite{kim2022infonerf} and depth based reprojection consistency~\cite{kwak2023geconerf, xu2022sinnerf,bortolon2022dvm,chen2022geoaug}.
    More recent works have explored other forms of regularizations.
    FreeNeRF~\cite{yang2023freenerf} anneals the frequency range of positional encoded NeRF inputs as the training progresses.
    MixNeRF~\cite{seo2023mixnerf} regularizes the NeRF by modeling the volume density along a ray as a mixture of Laplacian distributions.
    A recent work VDN-NeRF~\cite{zhu2023vdnnerf}, aims to resolve shape-radiance ambiguity, but is designed for training the NeRF with dense input views.
    However, our regularization is aimed at sparse input NeRF with as few as two views.


    \section{NeRF Preliminaries}\label{sec:nerf-preliminaries}
    We first provide a brief recap of the NeRF and describe the notation required for further sections.
    To render a pixel $\pixelq$, the NeRF shoots a corresponding ray into the scene and samples $N$ 3D points $\pointp_1, \pointp_2, \ldots, \pointp_N$, where $\pointp_1$ and $\pointp_N$ are the closest to and farthest from the camera, respectively.
    Two MLPs $\mlpf_1, \mlpf_2$ then predict view-independent volume density $\sigma_i$ and view-dependent color $\mathbf{c}_i$ as
    \begin{align}
        \sigma_i, \featureh_i = \mlpf_1 \left( \gamma(\pointp_i, 0, l_p) \right); \qquad \colorc_i = \mlpf_2 \left( \featureh_i, \gamma(\viewv, 0, l_v) \right),
    \end{align}
    where $\viewv$ is the viewing direction, $\featureh_i$ is a latent feature of $\pointp_i$ and
    \[\gamma(x, d_1, d_2) = [x, \sin(2^{d_1}x), \cos(2^{d_1}x), \ldots, \sin(2^{d_2-1}x), \cos(2^{d_2-1}x)]\]
    is the positional encoding.
    $l_p$ and $l_v$ are the highest positional encoding frequencies for $\pointp_i$ and $\viewv$ respectively.
    Volume rendering is then applied along every ray to obtain the color for each pixel as
    $\colorc = \sum_{i=1}^{N} w_i \colorc_i,$
    where the weights $w_i$ are computed as
    \begin{align}
        w_i = \exp \left( -\sum_{j=1}^{i-1} \delta_j \sigma_j \right) \cdot \left( 1 - \exp \left( -\delta_i \sigma_i \right) \right),
    \end{align}
    and $\delta_i$ is the distance between $\pointp_i$ and $\pointp_{i+1}$.
    The expected ray termination length is computed as $z = \sum_{i=1}^{N} w_i z_i$, where $z_i$ is the depth of $\pointp_i$.
    $z$ is typically also used as the depth of the pixel $\pixelq$~\cite{deng2022dsnerf}.
    $\mlpf_1$ and $\mlpf_2$ are trained using the mean squared error loss (MSE) as $\losscolor = \Vert \colorc - \hat{\colorc} \Vert^2,$ where $\hat{\colorc}$ is the true color of $\pixelq$.

    NeRF circumvents the need for the dense sampling of 3D points by employing two sets of MLPs, a coarse NeRF and a fine NeRF, both trained using $\losscolor$.
    The coarse NeRF is trained with a coarse stratified sampling, and the fine NeRF with dense sampling around object surfaces, where object surfaces are coarsely localized based on the predictions of the coarse NeRF.

    \begin{figure*}
        \includegraphics[width=0.95\linewidth]{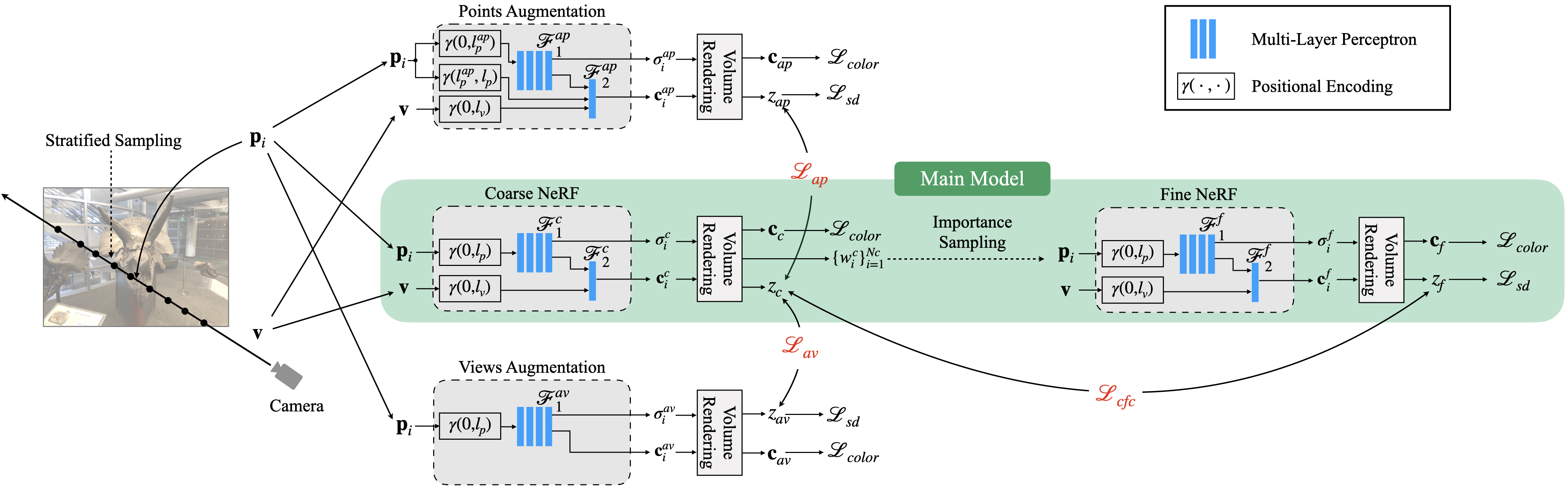}
        \caption{Architecture of SimpleNeRF.
        We train two augmented NeRF models in tandem with the NeRF to obtain depth supervision.
        In points augmentation, we reduce the positional encoding frequencies input to $\mlpf_1$ and concatenate them to the input of $\mlpf_2$.
        For views augmentation, we ask $\mlpf_1$ to output both volume density and color based on position alone.
        We add depth supervision losses \textcolor{red}{$\lossap$} and \textcolor{red}{$\lossav$} between the coarse NeRFs of the main and augmented models and a consistency loss \textcolor{red}{$\losscfc$} between the coarse and fine NeRFs of the main model.
        During inference, only the Main Model is employed.
        }
        \label{fig:model-architecture}
    \end{figure*}

    \section{Method}\label{sec:method}
    Our key idea is to employ simpler NeRF models to obtain better depth supervision in certain regions of the scene.
    We describe our regularizations based on the simpler solutions in \cref{subsec:simpler-solution-regularization}.
    We explain our approach to selecting reliable depth estimates for supervision in \cref{subsec:reliable-depth-estimates}.
    Finally, \cref{subsec:hierarchical-sampling} describes our solution to mitigate the sub-optimal utilization of hierarchical sampling.
    \cref{fig:model-architecture} shows the overall architecture of our model.

    \subsection{Regularization with Simpler Solutions}\label{subsec:simpler-solution-regularization}
    
    Our regularization consists of simplifying the NeRF model with respect to the positional encoding and view-dependent radiance capabilities to obtain better depth supervision. 
    Both positional encoding and view-dependent radiance are elements designed to increase the capability of the NeRF to explain complex phenomena.
    For example, the former helps in learning thin, repetitive objects against a farther background and the latter for specular objects.
    However, when training with sparse views, the fewer constraints coupled with the higher capacity of the NeRF lead to solutions that overfit the observed images and learn implausible scene geometries.

    We rethink ways of employing positional encoding and view-dependent radiance such that the NeRF utilizes the higher capability only when needed.
    The challenge here is that it is not known apriori where one needs to employ the higher capability of the NeRF\@.
    Our solution here is to use the higher capability NeRF as the main model and employ lower capability NeRFs as augmentations to provide guidance on where to use simpler solutions.
    Since the scene geometry is mainly learned by the coarse NeRF, we add the augmentations only to the coarse NeRF.
    We employ two augmentations, one each for regularizing positional encoding and view-dependent radiance, which we describe in the following subsections.
    In the remainder of this paper, we refer to the two augmentations as points (\cref{subsubsec:points-augmentation}) and views (\cref{subsubsec:views-augmentation}) augmentations respectively.


    \begin{table*}
        \centering
        \caption{Quantitative results on LLFF dataset.}
        \resizebox{\linewidth}{!}{
            \begin{tabular}{l|ccc|ccc|ccc}
                \hline
                & \multicolumn{3}{c|}{2 views} & \multicolumn{3}{c|}{3 views} & \multicolumn{3}{c}{4 views} \\
                \textbf{Model} &
                \textbf{LPIPS \textdownarrow} & \textbf{SSIM \textuparrow} & \textbf{PSNR \textuparrow} &
                \textbf{LPIPS \textdownarrow} & \textbf{SSIM \textuparrow} & \textbf{PSNR \textuparrow} &
                \textbf{LPIPS \textdownarrow} & \textbf{SSIM \textuparrow} & \textbf{PSNR \textuparrow} \\
                \hline
                RegNeRF           & 0.3056 & 0.5712 & 18.52 & 0.2908 & 0.6334 & 20.22 & 0.2794 & 0.6645 & 21.32 \\
                FreeNeRF          & \textbf{0.2638} & 0.6322 & 19.52 & 0.2754 & 0.6583 & 20.93 & 0.2848 & 0.6764 & 21.91 \\
                DS-NeRF           & 0.3106 & 0.5862 & 18.24 & 0.3031 & 0.6321 & 20.20 & 0.2979 & 0.6582 & 21.23 \\
                DDP-NeRF          & 0.2851 & 0.6218 & 18.73 & 0.3250 & 0.6152 & 18.73 & 0.3042 & 0.6558 & 20.17 \\
                ViP-NeRF          & 0.2768 & 0.6225 & 18.61 & 0.2798 & 0.6548 & 20.54 & 0.2854 & 0.6675 & 20.75 \\  
                SimpleNeRF        & 0.2688 & \textbf{0.6501} & \textbf{19.57} & \textbf{0.2559} & \textbf{0.6940} & \textbf{21.37} & \textbf{0.2633} & \textbf{0.7016} & \textbf{21.99} \\
                \hline
            \end{tabular}
        }
        \label{tab:quantitative-llff}
    \end{table*}
    
    \subsubsection{Undesirable Depth Discontinuities}\label{subsubsec:points-augmentation}
    The positional encoding maps two nearby points in $\mathbb{R}^3$ to two farther away points in $\mathbb{R}^{3(2l_p+1)}$ allowing the NeRF to learn sharp discontinuities between the two points in $\mathbb{R}^3$ as a smooth function in $\mathbb{R}^{3(2l_p+1)}$.
    However, this is mainly required at depth edges, and most regions of natural scenes are typically smooth in depth.
    With sparse input views, the positional encoding causes the NeRF to learn depth discontinuities even in smooth regions due to incorrect matching of pixels between the input views.
    This gives rise to ``floaters''~\cite{barron2022mipnerf360} where a part of an object is broken away from it and ``floats'' freely in space.
    We reduce the depth discontinuities by reducing the highest positional encoding frequency for $\pointp_i$ to $l_p^{\text{ap}} < l_p$ as
    \begin{align}
        \sigma_i, \mathbf{h}_i = \mathcal{F}^{\text{ap}}_1 (\gamma(\pointp_i, 0, l_p^{\text{ap}})), \label{eq:points-augmentation-mlp-f1}
    \end{align}
    where $\mlpf^{\text{ap}}_1$ is the MLP of the augmented model.
    The main model is more accurate where depth discontinuities are required and the augmented model is more accurate where discontinuities are not required.
    We determine the above using a ternary mask $m_{\text{ap}}$ as we explain in \cref{subsec:reliable-depth-estimates}.
    We supervise the depth predicted by the main NeRF using that of the augmented model, for those pixels for which the depth estimated by the augmented model is reliable, by setting $m_{\text{ap}} = 1$.
    Similarly, we also determine the pixels for which the depth estimated by the main model is more accurate than that of the augmented model, where we set $m_{\text{ap}} = -1$.
    In such pixels, we supervise the augmented model with the depth estimated by the main model, which helps the augmented model to improve further and provide better depth supervision for the main model.
    For pixels where the depth estimated by both the main and augmented models are unreliable, we set $m_{\text{ap}} = 0$.
    If $z_c$ and $z_{\text{ap}}$ are the depths estimated by the coarse NeRF of the main and the augmented models respectively, we impose the depth supervision as
    \begin{align}
        \lossap = \indicatorfunc_{\{m_{\text{ap}} = 1\}}  \odot \Vert z_c - \Tau(z_{\text{ap}}) \Vert^2 + \indicatorfunc_{\{m_{\text{ap}} = -1\}} \odot \Vert \Tau(z_c) - z_{\text{ap}} \Vert^2,
        \label{eq:loss-points-augmentation}
    \end{align}
    where $\odot$ denotes element-wise product, $\indicatorfunc$ is the indicator function and $\Tau$ is the stop-gradient operator.

    Since color tends to have more discontinuities than depth in regions such as textures, we include the remaining high-frequency positional encoding components of $\pointp_i$ in the input for $\mathcal{F}_2$ as
    \begin{align}
        \mathbf{c}_i = \mathcal{F}^{\text{ap}}_2 (\mathbf{h}_i, \gamma(\pointp_i, l_p^{\text{ap}}, l_p), \gamma(\mathbf{v}_i, 0, l_v)). \label{eq:points-augmentation-mlp-f2}
    \end{align}
    Note that $\featureh_i$ already includes the low-frequency positional encoding components of $\pointp_i$.

    \subsubsection{Shape-Radiance Ambiguity}\label{subsubsec:views-augmentation}
    The ability of the NeRF to predict view-dependent radiance helps it learn non-Lambertian surfaces.
    With fewer images, the NeRF can simply learn any random geometry and change the color of 3D points in accordance with the input viewpoint to explain away the observed images~\cite{zhang2020nerfpp}
    To bias the NeRF against this, we disable the view-dependent radiance in the second augmented NeRF model to output color based on $\pointp_i$ alone.
    If $z_{\text{av}}$ is the depth estimated by the augmented model, we impose the depth supervision as
    \begin{align}
        \lossav = \indicatorfunc_{\{m_{\text{av}} = 1\}}  \odot \Vert z_c - \Tau(z_{\text{av}}) \Vert^2 + \indicatorfunc_{\{m_{\text{av}} = -1\}} \odot \Vert \Tau(z_c) - z_{\text{av}} \Vert^2,
        \label{eq:loss-views-augmentation}
    \end{align}
    where $m_{\text{av}}$ is a ternary mask indicating where the depths estimated by the augmented and the main model are reliable.
    We note that while the augmented model is more accurate in Lambertian regions, the main model is better equipped to handle specular objects.

    \subsection{Determining Reliable Depth Estimates}\label{subsec:reliable-depth-estimates}


    We follow similar procedures to determine the masks $m_{\text{ap}}$ and $m_{\text{av}}$ and hence explain the mask computation using a generic variable $m_a$ to denote either of the masks.
    Given the depths $z_c$ and $z_a$ estimated by the main and augmented models respectively for pixel $\pixelq$, we reproject a $k \times k$ patch around $\pixelq$ to the nearest training view using both $z_c$ and $z_a$.
    We compute the MSE in intensities between the reprojected patch and the corresponding patch in the training image and choose the depth corresponding to lower MSE as the reliable depth.
    To filter out the cases where both the main and augmented models predict incorrect depth, we define a threshold $e_\tau$ and mark the depth to be reliable if its corresponding MSE is also less than $e_\tau$.
    If $e_c$ and $e_a$ are the reprojection MSE corresponding to $z_c$ and $z_a$ respectively, we compute the mask as
    \begin{align}
        m_a =
        \begin{cases}
            1 \qquad\ &\text{ if } (e_a \le e_c) \ \text{and} \ (e_a \le e_\tau)\\
            -1 \qquad\ &\text{ if } (e_c < e_a) \ \text{and} \ (e_c \le e_\tau)\\
            0 &\text{ otherwise }.
        \end{cases}
        \label{eq:mask-stable-sample}
    \end{align}
    For specular regions, the reprojected patches may not match in intensities leading to $m_a$ being zero.
    This implies supervision for fewer pixels and not supervision with incorrect depth estimates.
    We note that although both the views augmentation and the depth reliability estimation work only in the non-specular regions, there is no redundancy in the model.
    The views augmentation model may still make errors in the non-specular regions, and hence it is necessary to determine the reliability of its depth estimates.

    \subsection{Hierarchical Sampling}\label{subsec:hierarchical-sampling}
    Since multiple solutions can explain the observed images in the few-shot setting, the coarse and fine MLP may converge to different depth estimates for a given pixel as shown in \cref{fig:teaser-hierarchical-sampling2}.
    Thus, dense sampling may not be employed around the region where fine NeRF predicts the object surface, which is equivalent to using only the coarse sampling for the fine NeRF\@.
    This can lead to blur in rendered images as seen in \cref{fig:teaser-hierarchical-sampling1}.
    To prevent such inconsistencies, we drive the two NeRFs to be consistent in their solutions by imposing an MSE loss between the depths predicted by the two NeRFs.
    If $z_c$ and $z_f$ are the depths estimated by the coarse and fine NeRFs respectively, we define the coarse-fine consistency loss as
    \begin{align}
        \losscfc = \indicatorfunc_{\{m_{\text{cfc}} = 1\}}  \odot \Vert z_c - \Tau(z_f) \Vert^2 + \indicatorfunc_{\{m_{\text{cfc}} = -1\}} \odot \Vert \Tau(z_c) - z_f \Vert^2,
        \label{eq:loss-coarse-fine-consistency}
    \end{align}
    where the mask $m_{\text{cfc}}$ is determined as described in \cref{subsec:reliable-depth-estimates}.


    \begin{table*}
        \centering
        \caption{Quantitative results on RealEstate-10K dataset.}
        \resizebox{\linewidth}{!}{
            \begin{tabular}{l|ccc|ccc|ccc}
                \hline
                & \multicolumn{3}{c|}{2 views} & \multicolumn{3}{c|}{3 views} & \multicolumn{3}{c}{4 views} \\
                \textbf{Model} &
                \textbf{LPIPS \textdownarrow} & \textbf{SSIM \textuparrow} & \textbf{PSNR \textuparrow} &
                \textbf{LPIPS \textdownarrow} & \textbf{SSIM \textuparrow} & \textbf{PSNR \textuparrow} &
                \textbf{LPIPS \textdownarrow} & \textbf{SSIM \textuparrow} & \textbf{PSNR \textuparrow} \\
                \hline
                RegNeRF     & 0.4129 & 0.5916 & 17.14 & 0.4171 & 0.6132 & 17.86 & 0.4316 & 0.6257 & 18.34 \\
                FreeNeRF    & 0.5036 & 0.5354 & 14.70 & 0.4635 & 0.5708 & 15.26 & 0.5226 & 0.6027 & 16.31 \\
                DS-NeRF     & 0.2709 & 0.7983 & 26.26 & 0.2893 & 0.8004 & 26.50 & 0.3103 & 0.7999 & 26.65 \\
                DDP-NeRF    & 0.1290 & 0.8640 & 27.79 & 0.1518 & 0.8587 & 26.67 & 0.1563 & 0.8617 & 27.07 \\
                ViP-NeRF    & 0.0687 & 0.8889 & 32.32 & 0.0758 & 0.8967 & 31.93 & 0.0892 & 0.8968 & 31.95 \\  
                SimpleNeRF  & \textbf{0.0635} & \textbf{0.8942} & \textbf{33.10} & \textbf{0.0726} & \textbf{0.8984} & \textbf{33.21} & \textbf{0.0847} & \textbf{0.8987} & \textbf{32.88} \\  
                \hline
            \end{tabular}
        }
        \label{tab:quantitative-realestate}
    \end{table*}

    \subsection{Overall Loss}\label{subsec:overall-loss}
    We impose the pixel color reconstruction loss on the main model and the augmented models as
    \begin{align}
        \losscolor = \Vert \colorc_c - \hat{\colorc} \Vert^2 + \Vert \colorc_f - \hat{\colorc} \Vert^2 + \Vert \colorc_{\text{ap}} - \hat{\colorc} \Vert^2 + \Vert \colorc_{\text{av}} - \hat{\colorc} \Vert^2, \label{eq:loss-color}
    \end{align}
    where the subscripts $c, f$, `ap', and `av' denote the outputs of the coarse NeRF, fine NeRF, points augmentation model, and the views augmentation model respectively.
    We also include the sparse depth loss on the models as,
    \begin{align}
        \losssd = \Vert z_f - \hat{z} \Vert^2 + \Vert z_{\text{ap}} - \hat{z} \Vert^2 + \Vert z_{\text{av}} - \hat{z} \Vert^2,
    \end{align}
    where $\hat{z}$ is the sparse depth given by the SfM model.
    We do not impose $\losssd$ on the coarse NeRF of the main model following \citet{deng2022dsnerf}.
    Our final loss is a combination of all the losses as
    \begin{align}
        \mathcal{L} = \lambda_1 \losscolor + \lambda_2 \losssd + \lambda_3 \lossap + \lambda_4 \lossav + \lambda_5 \losscfc
    \end{align}

    \section{Experiments}\label{sec:experiments}
    \begin{table}
        \centering
        \caption{Evaluation of depth estimated by different models with two input views.
        The reference depth is obtained using NeRF with dense input views.
        }
        \begin{tabular}{l|cc|cc}
            \hline
            & \multicolumn{2}{c|}{LLFF} & \multicolumn{2}{c}{RealEstate-10K} \\
            model &
            \textbf{MAE \textdownarrow} & \textbf{SROCC \textuparrow} &
            \textbf{MAE \textdownarrow} & \textbf{SROCC \textuparrow} \\
            \hline
            DS-NeRF    & 0.2074 & 0.7230 & 0.7164 & 0.6660 \\
            DDP-NeRF   & 0.2048 & 0.7480 & 0.4831 & 0.7921 \\
            ViP-NeRF   & 0.1999 & 0.7344 & 0.3856 & 0.8446 \\
            SimpleNeRF & \textbf{0.1420} & \textbf{0.8480} & \textbf{0.3269} & \textbf{0.9215} \\
            \hline
        \end{tabular}
        \label{tab:quantitative-depth}
    \end{table}
    \subsection{Evaluation Setup}\label{subsec:evaluation-setup}
    We compare the performance of sparse input NeRF models on LLFF~\cite{mildenhall2019llff} and RealEstate-10K~\cite{zhou2018stereomag} datasets with 2, 3, and 4 input views.
    We assume the camera parameters are known for the input images, since in applications such as robotics or extended reality, external sensors or pre-calibrated set of cameras may provide the camera poses.
    We follow prior work~\cite{somraj2023vipnerf} to choose the train and test images.
    We provide more details in the supplementary.

    We quantitatively evaluate the predicted frames from various models using peak signal to noise ratio (PSNR), structural similarity (SSIM)~\cite{wang2004image} and LPIPS~\cite{zhang2018unreasonable} measures.
    We employ depth mean absolute error (MAE) and spearman rank order correlation coefficient (SROCC) to evaluate the models on their ability to predict absolute and relative depth in novel views.
    We train a NeRF model with dense input views and use its depth predictions as pseudo ground truth.
    On the LLFF dataset, we normalize the predicted depths by the median ground truth depth, since the scenes have different depth ranges.
    Following RegNeRF~\cite{niemeyer2022regnerf}, we evaluate the predictions only in the regions of interest.
    We mask out the regions of test frames which are not visible in the train frames.
    To determine such regions, we use the depth estimated by a NeRF trained with dense input views and compute the visible region mask through reprojection error in depth.
    We provide more details on the mask computation in the supplementary along with the unmasked evaluation scores for reference.

    \begin{figure*}
        \centering
        \includegraphics[width=0.995\linewidth]{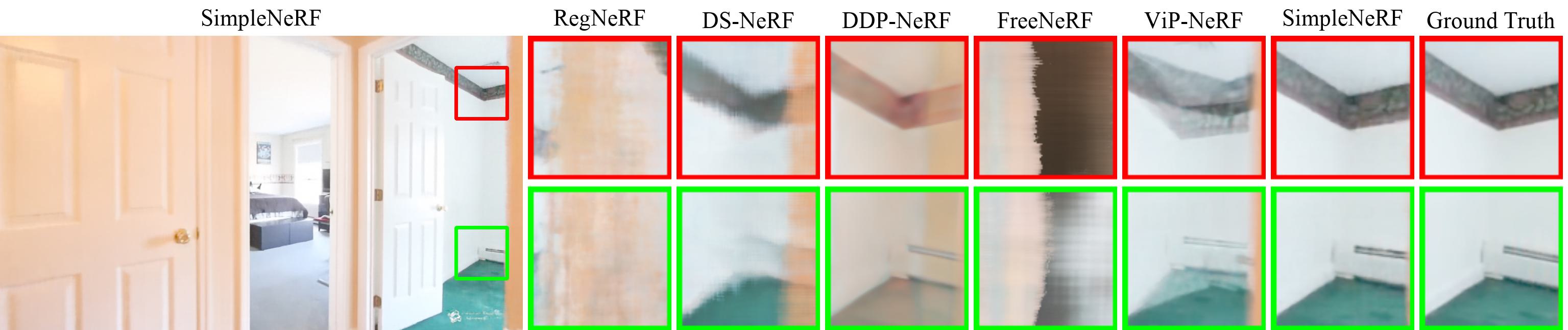}
        \caption{\textbf{Qualitative examples on RealEstate-10K dataset} with three input views.
        SimpleNeRF predictions are closest to the ground truth among all the models.
        In particular, DDP-NeRF predictions have a different shade of color and ViP-NeRF suffers from shape-radiance ambiguity creating ghosting artifacts.
        }
        \label{fig_qualitative-realestate02}
    \end{figure*}

    \subsection{Comparisons}\label{subsec:comparisons}
    We evaluate the performance of our model against various sparse input NeRF models on both datasets.
    We compare with DS-NeRF~\cite{deng2022dsnerf}, DDP-NeRF~\cite{roessle2022ddpnerf} and RegNeRF~\cite{niemeyer2022regnerf} which regularize the depth estimated by the NeRF\@.
    Further, we include two recent models, FreeNeRF~\cite{yang2023freenerf} and ViP-NeRF~\cite{somraj2023vipnerf}, among the comparisons.
    We train the models on both datasets using the codes provided by the respective authors.
    Implementation details of our model are provided in the supplement.

    \begin{figure}
        \centering
        \includegraphics[width=\linewidth]{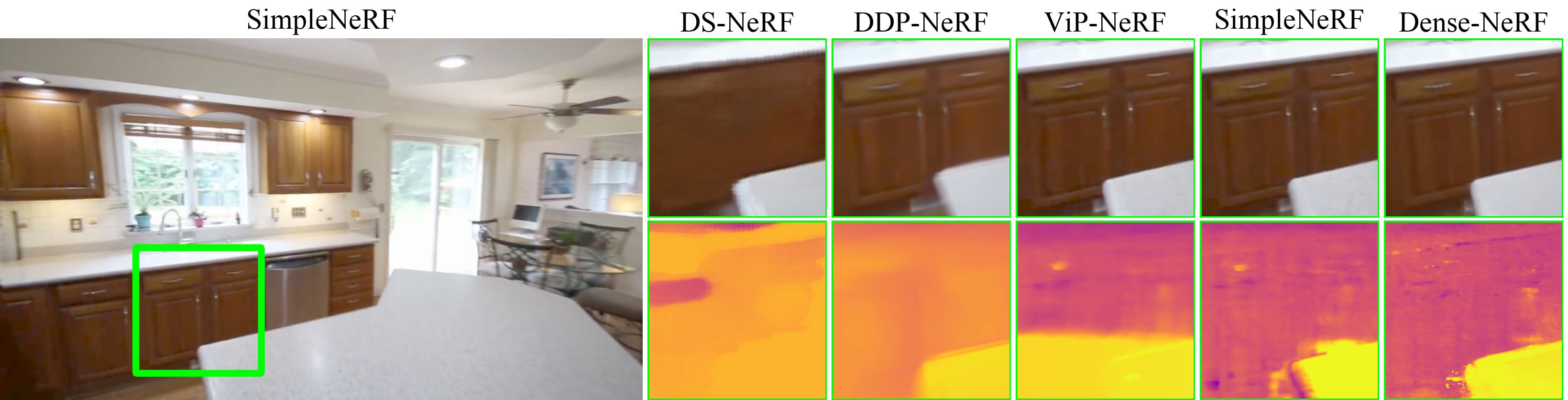}
        \includegraphics[width=\linewidth]{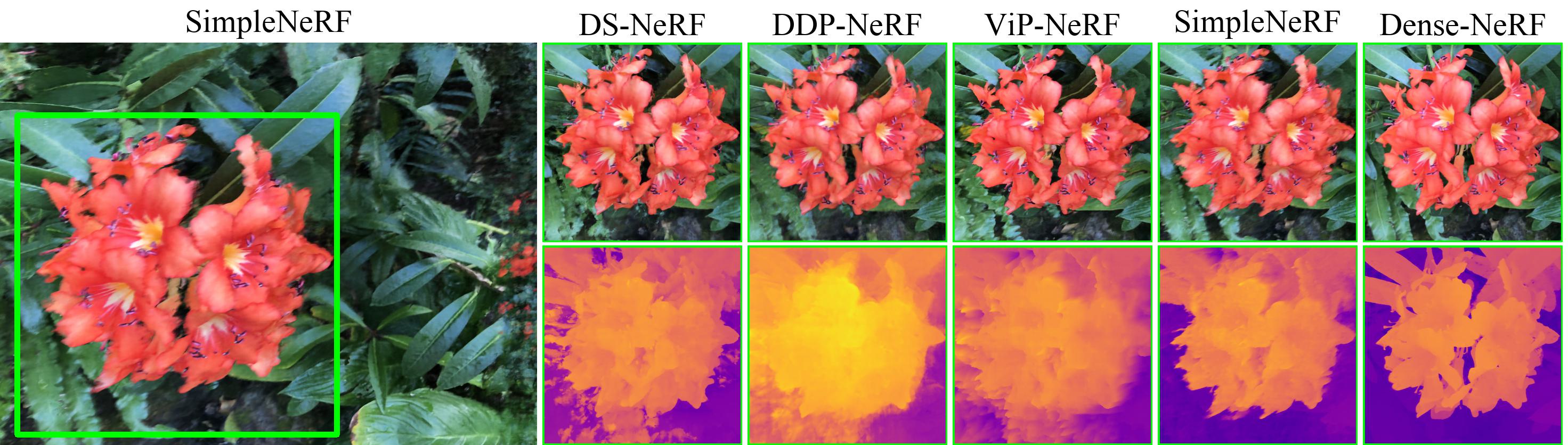}
        \caption{\textbf{Estimated depth maps} on RealEstate-10K and LLFF datasets with two input views.
        In both examples, the two rows show the predicted images and the depths respectively. 
        We find that SimpleNeRF is significantly better at estimating the scene depth.
        Also, DDP-NeRF synthesizes the left table edge at a different angle due to incorrect depth estimation.
        }
        \label{fig:qualitative-depth}
    \end{figure}

    \subsection{Results}\label{subsec:results}
    \cref{tab:quantitative-realestate,tab:quantitative-llff} show the view-synthesis performance of SimpleNeRF and other prior art on LLFF and RealEstate-10K datasets.
    We find that SimpleNeRF achieves state-of-the-art performance on both datasets in most cases.
    The higher performance of all the models on the RealEstate-10K dataset is perhaps due to the scenes being simpler.
    Hence, the performance improvement is also smaller as compared to the LLFF dataset.
    On RealEstate-10K, we observe that all the models struggle on one of the five scenes as compared to the other scenes.
    Excluding this scene, with two input views, SimpleNeRF improves SSIM over ViP-NeRF from 0.9596 to 0.9685, which we believe is a significant improvement at such high quality regime.
    In \cref{tab:quantitative-realestate}, we show the average performance on all five scenes and show the per-scene performance of various models in the supplementary.

\begin{figure*}
        \centering
        \includegraphics[width=\linewidth]{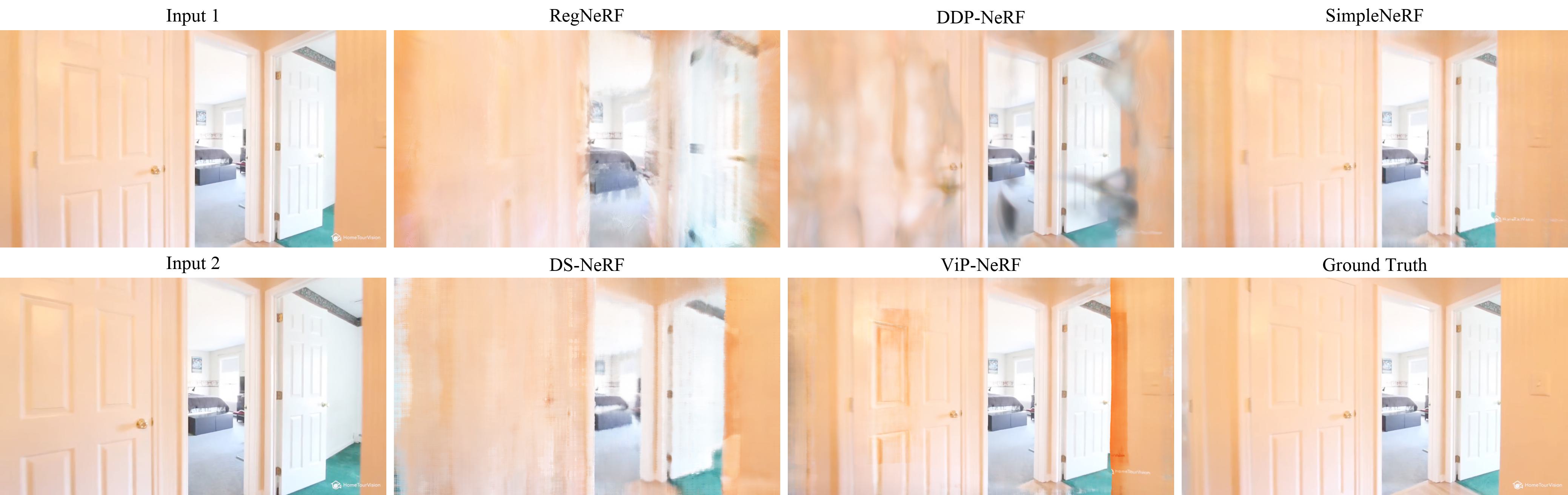}
        \caption{\textbf{Qualitative examples on the RealEstate-10K dataset with two input views.}
        While DDP-NeRF predictions contain blurred regions, ViP-NeRF predictions are color-saturated in certain regions of the door. 
        SimpleNeRF does not suffer from these distortions and synthesizes a clean frame.
        For reference, we also show the input images.
        }
        \label{fig:qualitative-realestate01}
    \end{figure*}

\begin{figure*}
        \centering
        \includegraphics[width=\linewidth]{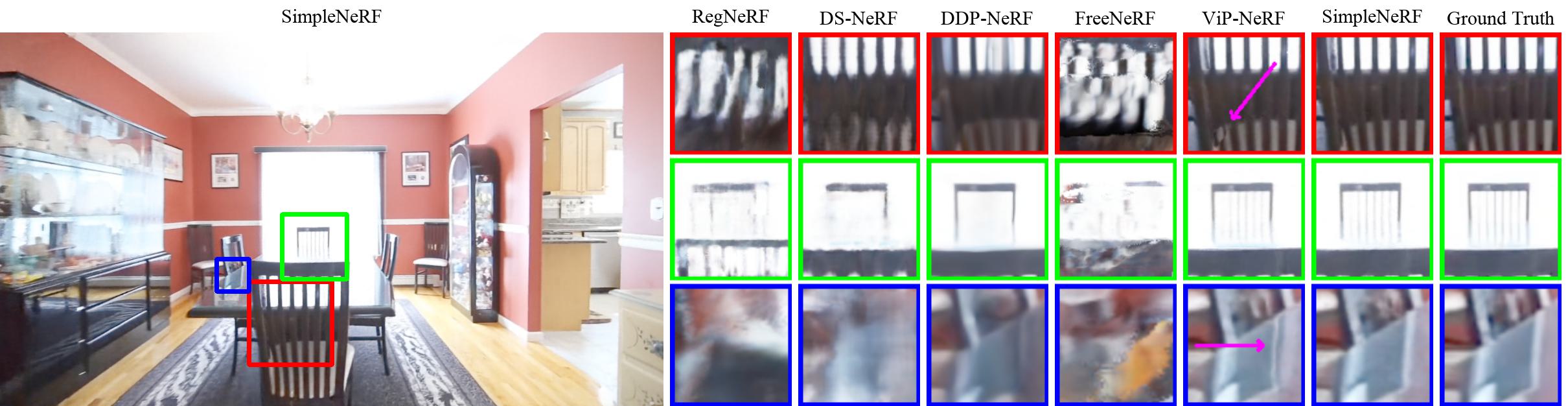}
        \caption{\textbf{Qualitative examples on the RealEstate-10K dataset with four input views.}
        We find that SimpleNeRF and ViP-NeRF perform the best among all the models.
        However, ViP-NeRF predictions contain minor distortions as pointed out by the magenta arrow, which is rectified by SimpleNeRF.
        }
        \label{fig:qualitative-realestate03}
    \end{figure*}

\begin{figure*}
        \centering
        \begin{subfigure}{0.54\linewidth}
            \centering
            \includegraphics[width=\linewidth]{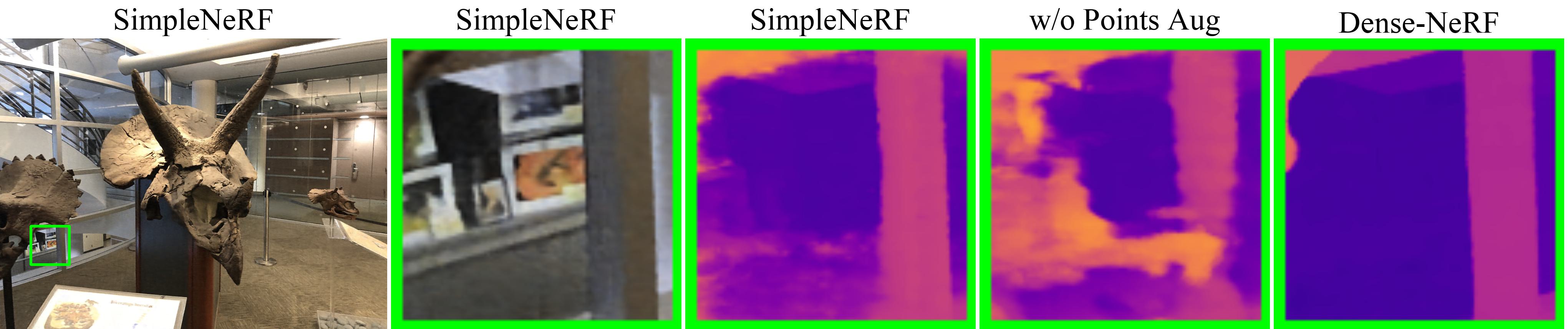}
            \caption{\textbf{Without Points Augmentation:}
            The ablated model introduces floaters that are significantly reduced by using the points augmentation model.
            }
            \label{fig:qualitative-ablation01a}
        \end{subfigure}
        \hfill
        \begin{subfigure}{0.44\linewidth}
            \centering
            \includegraphics[width=\linewidth]{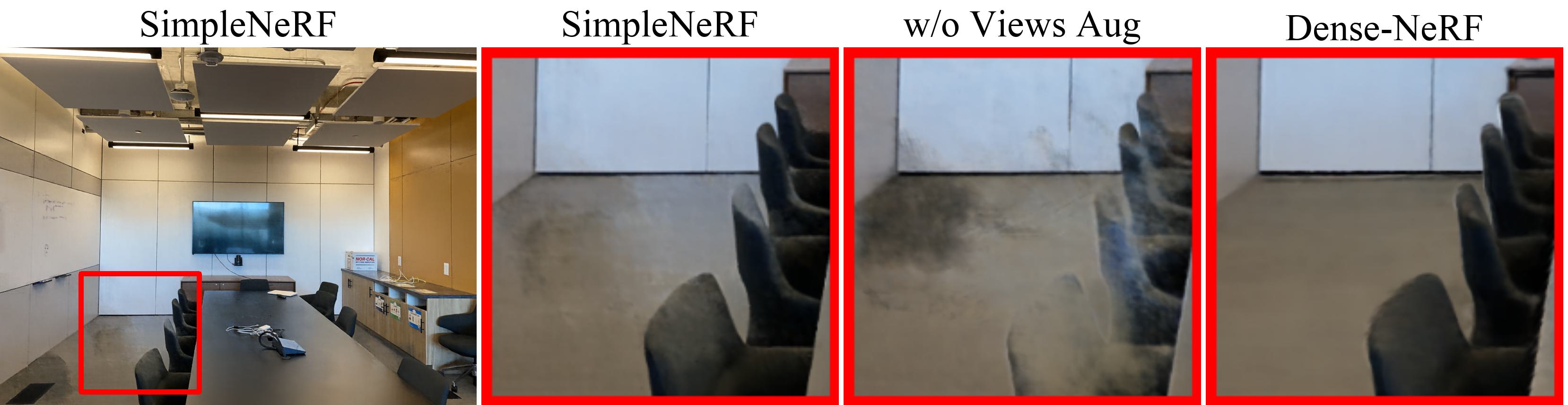}
            \caption{\textbf{Without Views Augmentation:}
            The ablated model suffers from shape-radiance ambiguity and produces ghosting artifacts.
            }
            \label{fig:qualitative-ablation01b}
        \end{subfigure}
        \begin{subfigure}{0.54\linewidth}
            \centering
            \includegraphics[width=\linewidth]{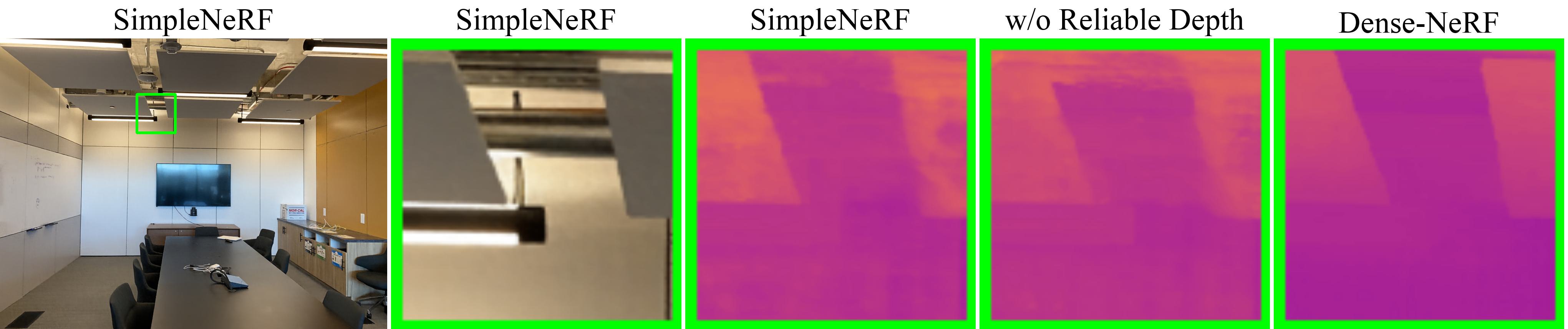}
            \caption{\textbf{Without Reliability of Depth Supervision:}
            The points augmentation model struggles to learn sharp depth discontinuities at true depth edges.
            Supervising the main model using such depths without determining their reliability causes the main model to learn incorrect depth.
            As a result, the ablated model fails to learn sharp depth discontinuities at certain regions.
            }
            \label{fig:qualitative-ablation01d}
        \end{subfigure}
        \hfill
        \begin{subfigure}{0.44\linewidth}
            \centering
            \includegraphics[width=\linewidth]{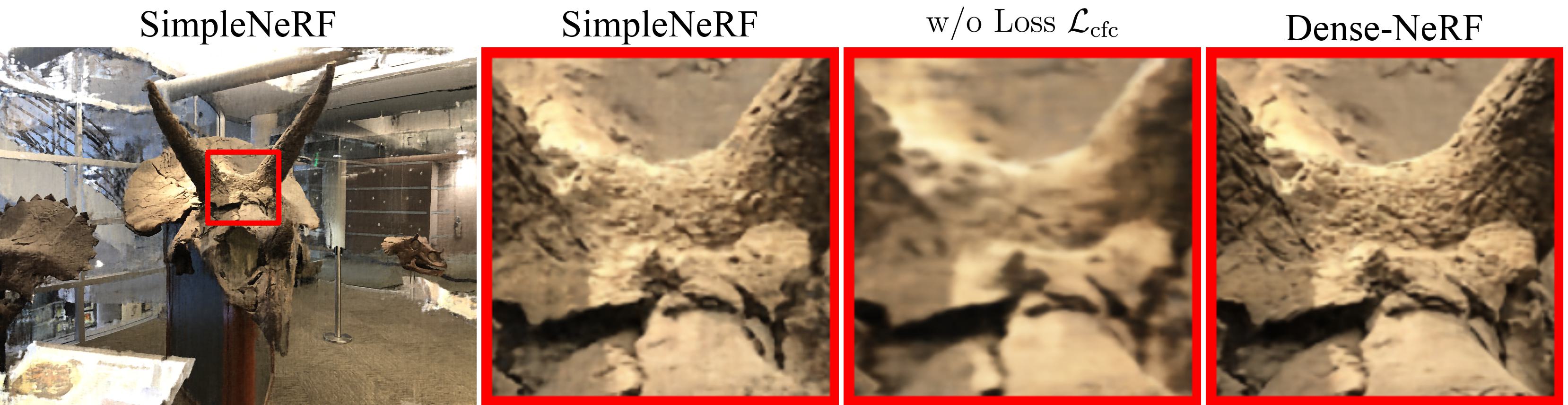}
            \caption{\textbf{Without Coarse-Fine Consistency:}
            We observe that while SimpleNeRF predictions are sharper, the ablated model without coarse-fine consistency loss, $\losscfc$ produces blurred renders.
            This is similar to \cref{fig:teaser-hierarchical-sampling2}, where we observe DS-NeRF also produce blurred renders.\\
            \ 
            }
            \label{fig:qualitative-ablation01c}
        \end{subfigure}
        \caption{Qualitative examples for ablated models on the LLFF dataset with two input views.
        We also show the outputs of the dense-input NeRF for reference.}
        \label{fig:qualitative-ablations01}
    \end{figure*}

\begin{figure*}
        \centering
        \includegraphics[width=\linewidth]{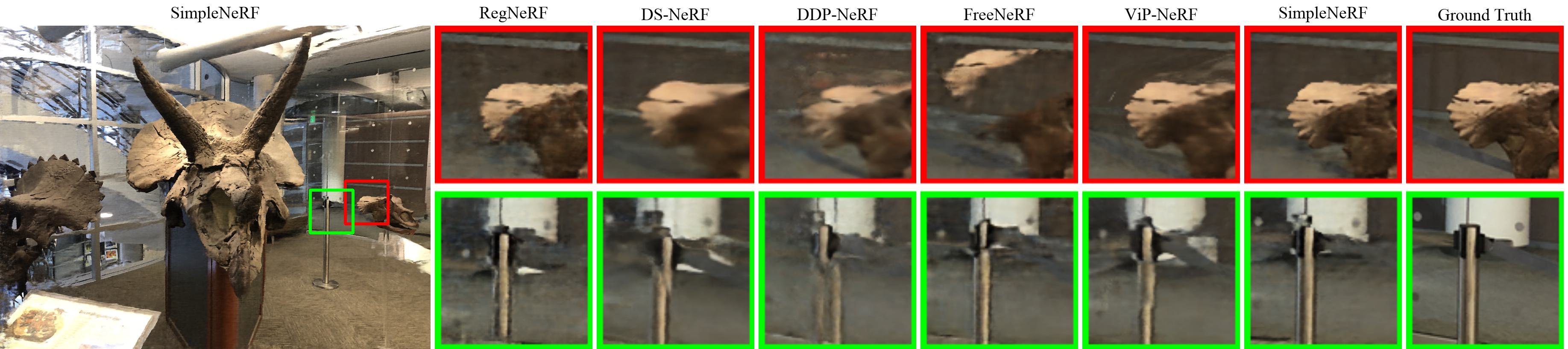}
        \caption{\textbf{Qualitative examples on the LLFF dataset with two input views.}
        DDP-NeRF and ViP-NeRF synthesize frames with broken objects in the second row and FreeNeRF breaks the object in the first row, due to incorrect depth estimations.
        SimpleNeRF produces sharper frames devoid of such artifacts.
        }
        \label{fig:qualitative-llff01}
    \end{figure*}

\begin{figure*}
        \centering
        \includegraphics[width=\linewidth]{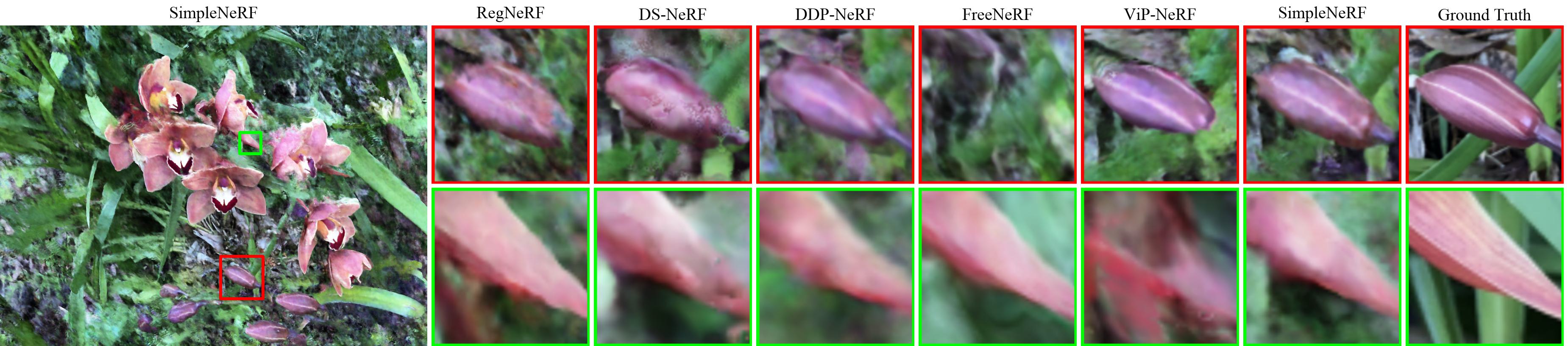}
        \caption{\textbf{Qualitative examples on the LLFF dataset with three input views.}
        In the first row, the orchid is displaced out of the cropped box in the FreeNeRF prediction, due to incorrect depth estimation.
        ViP-NeRF and RegNeRF fail to predict the complete orchid accurately and contain distortions at either ends.
        In the second row, ViP-NeRF prediction contains severe distortions.
        SimpleNeRF reconstructs the best among all the models in both examples.
        }
        \label{fig:qualitative-llff02}
    \end{figure*}

\begin{figure*}
        \centering
        \includegraphics[width=\linewidth]{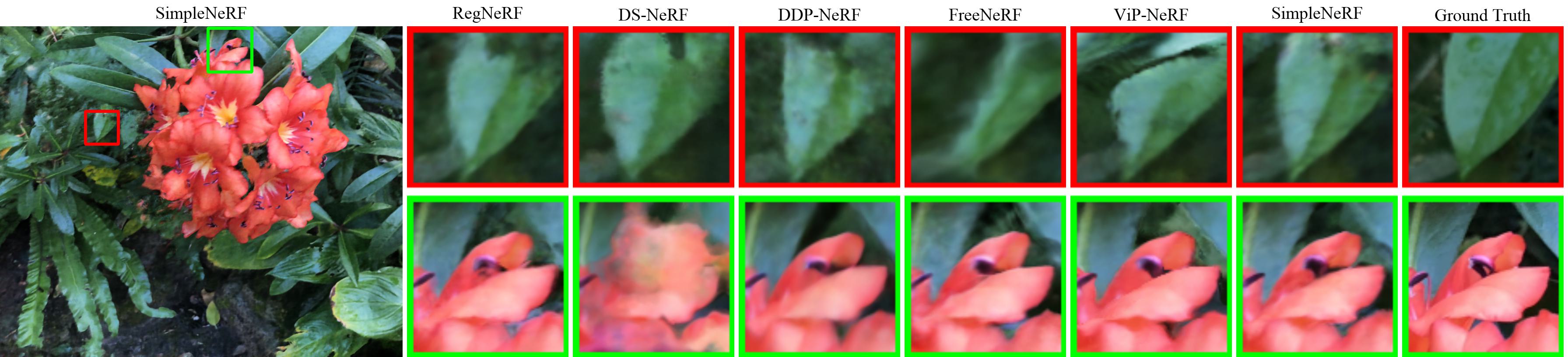}
        \caption{\textbf{Qualitative examples on the LLFF dataset with four input views.}
        In the first row, we find that ViP-NeRF, FreeNeRF, and DDP-NeRF struggle to reconstruct the shape of the leaf accurately.
        In the second row, DS-NeRF introduces floaters.
        SimpleNeRF does not suffer from such artifacts and reconstructs the shapes better.
        }
        \label{fig:qualitative-llff03}
    \end{figure*}

\begin{figure*}
        \centering
        \includegraphics[width=0.8\linewidth]{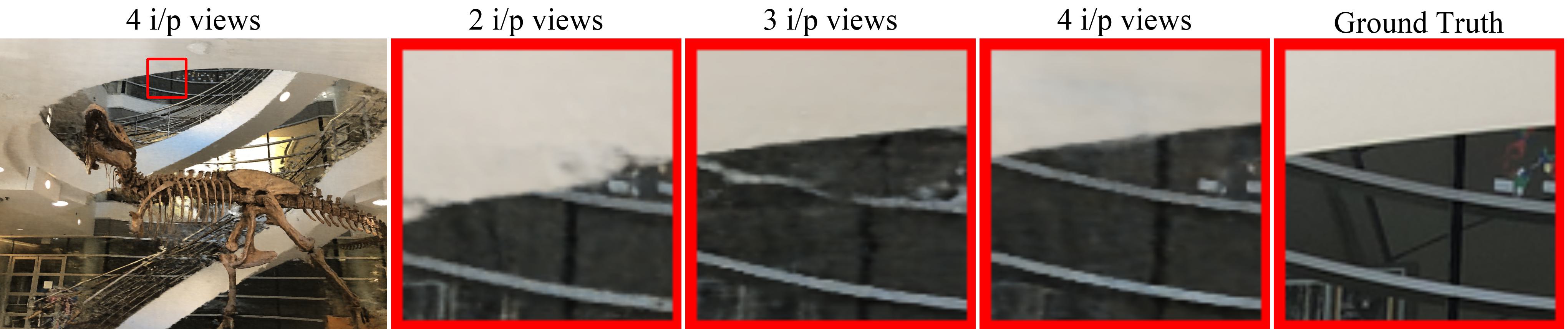}
        \caption{\textbf{Qualitative examples on the LLFF dataset with two, three, and four input views.}
        We observe errors in depth estimation with two input views causing a change in the position of the roof.
        While this is corrected with three input views, there are a few shape distortions in the metal rods.
        With four input views, even such distortions are corrected.
        }
        \label{fig:qualitative-llff04}
    \end{figure*}

    \cref{fig_qualitative-realestate02} shows predictions of various models on an example scene from the RealEstate-10K dataset, where we observe that SimpleNeRF is the best in reconstructing the novel view.
    \cref{fig:qualitative-realestate01,fig:qualitative-realestate03,fig:qualitative-ablations01,fig:qualitative-llff01,fig:qualitative-llff02,fig:qualitative-llff03,fig:qualitative-llff04} show more comparisons on both datasets.
    Further, SimpleNeRF improves significantly in estimating the depth of the scene as seen in ~\cref{tab:quantitative-ablations} and ~\cref{fig:qualitative-depth}.
    SimpleNeRF also performs significantly better in estimating relative depth, even outperforming ViP-NeRF which uses a prior based on relative depth.
    Estimating better geometry may be more crucial in downstream applications such as 3D scene editing.
    We provide video comparisons in the supplementary.

    \subsubsection{Ablations}\label{subsubsec:ablations}
    We test the importance of each of the components of our model components by disabling them one at a time.
    We disable the points and views augmentations and coarse-fine consistency loss individually.
    When disabling $\losscfc$, we additionally add augmentations to the fine NeRF since the knowledge learned by coarse NeRF may not propagate to the fine NeRF.
    We also analyze the need to supervise with only the reliable depth estimates by disabling the mask and stop-gradients in $\lossap, \lossav$, and $\losscfc$.
    \cref{tab:quantitative-ablations} shows a quantitative comparison between the ablated models.

    We observe that each of the components is crucial and disabling any of them leads to a drop in the performance.
    Further, using all the depths for supervision instead of only the reliable depths leads to a significant drop in performance.
    Finally, disabling $\losscfc$ also leads to a drop in performance in addition to increasing the training time by almost $2\times$ due to the inclusion of augmentations for the fine NeRF.

    We conduct two additional experiments to further analyze the impact of the augmentation we design.
    Firstly, we analyze if there is a need to design simpler augmentations by replacing our novel augmentations with identical replicas of the NeRF as augmentations.
    In the second experiment, we use a smaller network, by reducing the number of layers from eight to four, for the points augmentation instead of using fewer positional encoding frequencies.
    \cref{tab:quantitative-ablations} shows that in both the above cases, the performance drops to that of SimpleNeRF without the points augmentation or lower.
    In particular, reducing positional encoding frequencies has a higher impact than reducing the number of network layers perhaps because the network with fewer layers may still be capable of learning floaters on account of using all the positional encoding frequencies.
    In the supplement, we analyze how the performance of SimpleNeRF varies as $l_p^{\text{ap}}$ varies.

    \begin{table}
        \centering
        \setlength\tabcolsep{2pt}
        \caption{Ablation experiments on both datasets with two input views.}
        \begin{tabular}{l|cc|cc}
            \hline
            & \multicolumn{2}{c|}{RealEstate-10K} & \multicolumn{2}{c}{LLFF} \\
            model                & \textbf{LPIPS \textdownarrow} & \textbf{MAE \textdownarrow} & \textbf{LPIPS \textdownarrow} & \textbf{MAE \textdownarrow} \\
            \hline
            SimpleNeRF                  & \textbf{0.0635} & \textbf{0.33} & \textbf{0.2688} & \textbf{0.14} \\
            w/o points augmentation     & 0.0752 & 0.38 & 0.2832 & 0.15 \\
            w/o views augmentation      & 0.0790 & 0.39 & 0.2834 & 0.15 \\
            w/o coarse-fine consistency & 0.0740 & 0.42 & 0.3002 & 0.19 \\
            w/o reliable depth          & 0.0687 & 0.45 & 0.3020 & 0.22 \\
            \hline
            w/ identical augmentations  & 0.0777 & 0.40 & 0.2849 & 0.15 \\
            w/ smaller n/w as points aug   & 0.0740 & 0.38 & 0.2849 & 0.15 \\
            \hline
        \end{tabular}
        \label{tab:quantitative-ablations}
    \end{table}

    \subsection{Limitations}\label{subsec:limitations}
    Our approach to determining reliable depth estimates for supervision depends on the reprojection error, which may be high for specular objects even if the depth estimates are correct.
    It may be helpful to explore approaches to determine the reliability of depth estimates without employing the reprojection error.
    The shape of the reprojected patches to determine reliable depth estimates may change significantly if the input viewpoints are diverse.
    This can lead to incorrect mask estimation.
    Further, our model requires accurate camera poses of the sparse input images.
    Finally, the use of augmented models adds to computational and memory overhead during training by about 1.5 times.

    \section{Conclusion}\label{sec:conclusion}
    We address the problem of few-shot NeRF by obtaining depth supervision through simpler augmented models that are trained in tandem with the NeRF\@.
    We design two augmentations that learn simpler solutions and help the main NeRF model mitigate floater artifacts and shape-radiance ambiguity.
    By imposing a consistency loss between the coarse and fine NeRFs, we ensure better application of hierarchical sampling leading to sharper predictions.
    SimpleNeRF achieves state-of-the-art performance on two commonly used datasets in synthesizing novel views as well as estimating better scene geometry.
    In future, we plan to explore the role of simpler augmentations for newer models such as instant-ngp~\cite{muller2022instant} and related applications such as neural surfaces~\cite{yariv2021volume} to train them with sparse input views.

\begin{acks}
    This work was supported in part by a grant from Qualcomm.
    The first author was supported by the Prime Minister’s Research Fellowship awarded by the Ministry of Education, Government of India.
    The authors would also like to thank Shankhanil Mitra for valuable discussions.
\end{acks}

    \bibliographystyle{ACM-Reference-Format}
    \bibliography{SSLN}


\begin{thebibliography}{52}


\ifx \showCODEN    \undefined \def \showCODEN     #1{\unskip}     \fi
\ifx \showDOI      \undefined \def \showDOI       #1{#1}\fi
\ifx \showISBNx    \undefined \def \showISBNx     #1{\unskip}     \fi
\ifx \showISBNxiii \undefined \def \showISBNxiii  #1{\unskip}     \fi
\ifx \showISSN     \undefined \def \showISSN      #1{\unskip}     \fi
\ifx \showLCCN     \undefined \def \showLCCN      #1{\unskip}     \fi
\ifx \shownote     \undefined \def \shownote      #1{#1}          \fi
\ifx \showarticletitle \undefined \def \showarticletitle #1{#1}   \fi
\ifx \showURL      \undefined \def \showURL       {\relax}        \fi
\providecommand\bibfield[2]{#2}
\providecommand\bibinfo[2]{#2}
\providecommand\natexlab[1]{#1}
\providecommand\showeprint[2][]{arXiv:#2}

\bibitem[Barron et~al\mbox{.}(2022)]%
        {barron2022mipnerf360}
\bibfield{author}{\bibinfo{person}{Jonathan~T. Barron}, \bibinfo{person}{Ben
  Mildenhall}, \bibinfo{person}{Dor Verbin}, \bibinfo{person}{Pratul~P.
  Srinivasan}, {and} \bibinfo{person}{Peter Hedman}.}
  \bibinfo{year}{2022}\natexlab{}.
\newblock \showarticletitle{{Mip-NeRF} 360: Unbounded Anti-Aliased Neural
  Radiance Fields}. In \bibinfo{booktitle}{\emph{Proceedings of the IEEE/CVF
  Conference on Computer Vision and Pattern Recognition (CVPR)}}.
\newblock


\bibitem[Bian et~al\mbox{.}(2023)]%
        {bian2023nopenerf}
\bibfield{author}{\bibinfo{person}{Wenjing Bian}, \bibinfo{person}{Zirui Wang},
  \bibinfo{person}{Kejie Li}, \bibinfo{person}{Jia-Wang Bian}, {and}
  \bibinfo{person}{Victor~Adrian Prisacariu}.} \bibinfo{year}{2023}\natexlab{}.
\newblock \showarticletitle{{NoPe-NeRF}: Optimising Neural Radiance Field with
  No Pose Prior}.
\newblock  (\bibinfo{date}{June} \bibinfo{year}{2023}).
\newblock


\bibitem[Bortolon et~al\mbox{.}(2022)]%
        {bortolon2022dvm}
\bibfield{author}{\bibinfo{person}{Matteo Bortolon}, \bibinfo{person}{Alessio
  Del~Bue}, {and} \bibinfo{person}{Fabio Poiesi}.}
  \bibinfo{year}{2022}\natexlab{}.
\newblock \showarticletitle{Data augmentation for {NeRF}: a geometric
  consistent solution based on view morphing}.
\newblock \bibinfo{journal}{\emph{arXiv e-prints}}, Article
  \bibinfo{articleno}{arXiv:2210.04214} (\bibinfo{year}{2022}),
  \bibinfo{numpages}{arXiv:2210.04214}~pages.
\newblock
\showeprint[arxiv]{2210.04214}


\bibitem[Chen et~al\mbox{.}(2022b)]%
        {chen2022tensorf}
\bibfield{author}{\bibinfo{person}{Anpei Chen}, \bibinfo{person}{Zexiang Xu},
  \bibinfo{person}{Andreas Geiger}, \bibinfo{person}{Jingyi Yu}, {and}
  \bibinfo{person}{Hao Su}.} \bibinfo{year}{2022}\natexlab{b}.
\newblock \showarticletitle{{TensoRF}: Tensorial Radiance Fields}. In
  \bibinfo{booktitle}{\emph{Proceedings of the European Conference on Computer
  Vision (ECCV)}}.
\newblock


\bibitem[Chen et~al\mbox{.}(2021)]%
        {chen2021mvsnerf}
\bibfield{author}{\bibinfo{person}{Anpei Chen}, \bibinfo{person}{Zexiang Xu},
  \bibinfo{person}{Fuqiang Zhao}, \bibinfo{person}{Xiaoshuai Zhang},
  \bibinfo{person}{Fanbo Xiang}, \bibinfo{person}{Jingyi Yu}, {and}
  \bibinfo{person}{Hao Su}.} \bibinfo{year}{2021}\natexlab{}.
\newblock \showarticletitle{{MVSNeRF}: Fast Generalizable Radiance Field
  Reconstruction from Multi-View Stereo}.
\newblock \bibinfo{journal}{\emph{arXiv e-prints}}, Article
  \bibinfo{articleno}{arXiv:2103.15595} (\bibinfo{date}{March}
  \bibinfo{year}{2021}), \bibinfo{numpages}{arXiv:2103.15595}~pages.
\newblock
\showeprint[arxiv]{2103.15595}


\bibitem[Chen et~al\mbox{.}(2022a)]%
        {chen2022geoaug}
\bibfield{author}{\bibinfo{person}{Di Chen}, \bibinfo{person}{Yu Liu},
  \bibinfo{person}{Lianghua Huang}, \bibinfo{person}{Bin Wang}, {and}
  \bibinfo{person}{Pan Pan}.} \bibinfo{year}{2022}\natexlab{a}.
\newblock \showarticletitle{{GeoAug}: Data Augmentation for Few-Shot {NeRF}
  with Geometry Constraints}. In \bibinfo{booktitle}{\emph{Proceedings of the
  European Conference on Computer Vision (ECCV)}}.
\newblock


\bibitem[Chen and Williams(1993)]%
        {chen1993view}
\bibfield{author}{\bibinfo{person}{Shenchang~Eric Chen} {and}
  \bibinfo{person}{Lance Williams}.} \bibinfo{year}{1993}\natexlab{}.
\newblock \showarticletitle{View Interpolation for Image Synthesis}. In
  \bibinfo{booktitle}{\emph{Proceedings of the Computer Graphics and
  Interactive Techniques (SIGGRAPH)}}.
\newblock
\urldef\tempurl%
\url{https://doi.org/10.1145/166117.166153}
\showDOI{\tempurl}


\bibitem[Chen et~al\mbox{.}(2023)]%
        {chen2023explicit}
\bibfield{author}{\bibinfo{person}{Yuedong Chen}, \bibinfo{person}{Haofei Xu},
  \bibinfo{person}{Qianyi Wu}, \bibinfo{person}{Chuanxia Zheng},
  \bibinfo{person}{Tat-Jen Cham}, {and} \bibinfo{person}{Jianfei Cai}.}
  \bibinfo{year}{2023}\natexlab{}.
\newblock \showarticletitle{Explicit Correspondence Matching for Generalizable
  Neural Radiance Fields}.
\newblock \bibinfo{journal}{\emph{arXiv e-prints}}, Article
  \bibinfo{articleno}{arXiv:2304.12294} (\bibinfo{year}{2023}),
  \bibinfo{numpages}{arXiv:2304.12294}~pages.
\newblock
\showeprint[arxiv]{2304.12294}


\bibitem[Chibane et~al\mbox{.}(2021)]%
        {chibane2021srf}
\bibfield{author}{\bibinfo{person}{Julian Chibane}, \bibinfo{person}{Aayush
  Bansal}, \bibinfo{person}{Verica Lazova}, {and} \bibinfo{person}{Gerard
  Pons-Moll}.} \bibinfo{year}{2021}\natexlab{}.
\newblock \showarticletitle{Stereo Radiance Fields ({SRF}): Learning View
  Synthesis for Sparse Views of Novel Scenes}. In
  \bibinfo{booktitle}{\emph{Proceedings of the IEEE/CVF Conference on Computer
  Vision and Pattern Recognition (CVPR)}}.
\newblock


\bibitem[Cho et~al\mbox{.}(2017)]%
        {cho2017hole}
\bibfield{author}{\bibinfo{person}{Jea-Hyung Cho}, \bibinfo{person}{Wonseok
  Song}, \bibinfo{person}{Hyuk Choi}, {and} \bibinfo{person}{Taejeong Kim}.}
  \bibinfo{year}{2017}\natexlab{}.
\newblock \showarticletitle{Hole Filling Method for Depth Image Based Rendering
  Based on Boundary Decision}.
\newblock \bibinfo{journal}{\emph{IEEE Signal Processing Letters (SPL)}}
  \bibinfo{volume}{24}, \bibinfo{number}{3} (\bibinfo{year}{2017}),
  \bibinfo{pages}{329--333}.
\newblock


\bibitem[Deng et~al\mbox{.}(2022b)]%
        {deng2022dsnerf}
\bibfield{author}{\bibinfo{person}{Kangle Deng}, \bibinfo{person}{Andrew Liu},
  \bibinfo{person}{Jun-Yan Zhu}, {and} \bibinfo{person}{Deva Ramanan}.}
  \bibinfo{year}{2022}\natexlab{b}.
\newblock \showarticletitle{Depth-Supervised {NeRF}: Fewer Views and Faster
  Training for Free}. In \bibinfo{booktitle}{\emph{Proceedings of the IEEE/CVF
  Conference on Computer Vision and Pattern Recognition (CVPR)}}.
\newblock


\bibitem[Deng et~al\mbox{.}(2022a)]%
        {deng2022fovnerf}
\bibfield{author}{\bibinfo{person}{Nianchen Deng}, \bibinfo{person}{Zhenyi He},
  \bibinfo{person}{Jiannan Ye}, \bibinfo{person}{Budmonde Duinkharjav},
  \bibinfo{person}{Praneeth Chakravarthula}, \bibinfo{person}{Xubo Yang}, {and}
  \bibinfo{person}{Qi Sun}.} \bibinfo{year}{2022}\natexlab{a}.
\newblock \showarticletitle{{FoV-NeRF}: Foveated Neural Radiance Fields for
  Virtual Reality}.
\newblock \bibinfo{journal}{\emph{IEEE Transactions on Visualization and
  Computer Graphics (TVCG)}} \bibinfo{volume}{28}, \bibinfo{number}{11}
  (\bibinfo{year}{2022}), \bibinfo{pages}{3854--3864}.
\newblock
\urldef\tempurl%
\url{https://doi.org/10.1109/TVCG.2022.3203102}
\showDOI{\tempurl}


\bibitem[Gortler et~al\mbox{.}(1996)]%
        {gortler1996lumigraph}
\bibfield{author}{\bibinfo{person}{Steven~J. Gortler}, \bibinfo{person}{Radek
  Grzeszczuk}, \bibinfo{person}{Richard Szeliski}, {and}
  \bibinfo{person}{Michael~F. Cohen}.} \bibinfo{year}{1996}\natexlab{}.
\newblock \showarticletitle{The Lumigraph}. In
  \bibinfo{booktitle}{\emph{Proceedings of the Computer Graphics and
  Interactive Techniques (SIGGRAPH)}}.
\newblock
\urldef\tempurl%
\url{https://doi.org/10.1145/237170.237200}
\showDOI{\tempurl}


\bibitem[Jain et~al\mbox{.}(2021)]%
        {jain2021dietnerf}
\bibfield{author}{\bibinfo{person}{Ajay Jain}, \bibinfo{person}{Matthew
  Tancik}, {and} \bibinfo{person}{Pieter Abbeel}.}
  \bibinfo{year}{2021}\natexlab{}.
\newblock \showarticletitle{Putting {NeRF} on a Diet: Semantically Consistent
  Few-Shot View Synthesis}. In \bibinfo{booktitle}{\emph{Proceedings of the
  IEEE/CVF International Conference on Computer Vision (ICCV)}}.
\newblock


\bibitem[Johari et~al\mbox{.}(2022)]%
        {johari2022geonerf}
\bibfield{author}{\bibinfo{person}{Mohammad~Mahdi Johari},
  \bibinfo{person}{Yann Lepoittevin}, {and} \bibinfo{person}{Fran\c{c}ois
  Fleuret}.} \bibinfo{year}{2022}\natexlab{}.
\newblock \showarticletitle{GeoNeRF: Generalizing NeRF With Geometry Priors}.
  In \bibinfo{booktitle}{\emph{Proceedings of the IEEE/CVF Conference on
  Computer Vision and Pattern Recognition (CVPR)}}.
\newblock


\bibitem[Kanchana et~al\mbox{.}(2022)]%
        {kanchana2022ivp}
\bibfield{author}{\bibinfo{person}{Vijayalakshmi Kanchana},
  \bibinfo{person}{Nagabhushan Somraj}, \bibinfo{person}{Suraj Yadwad}, {and}
  \bibinfo{person}{Rajiv Soundararajan}.} \bibinfo{year}{2022}\natexlab{}.
\newblock \showarticletitle{Revealing Disocclusions in Temporal View Synthesis
  through Infilling Vector Prediction}. In
  \bibinfo{booktitle}{\emph{Proceedings of the IEEE Winter Conference on
  Applications of Computer Vision (WACV)}}.
\newblock


\bibitem[Ke et~al\mbox{.}(2019)]%
        {ke2019dualstudent}
\bibfield{author}{\bibinfo{person}{Zhanghan Ke}, \bibinfo{person}{Daoye Wang},
  \bibinfo{person}{Qiong Yan}, \bibinfo{person}{Jimmy Ren}, {and}
  \bibinfo{person}{Rynson~W.H. Lau}.} \bibinfo{year}{2019}\natexlab{}.
\newblock \showarticletitle{Dual Student: Breaking the Limits of the Teacher in
  Semi-Supervised Learning}. In \bibinfo{booktitle}{\emph{Proceedings of the
  IEEE/CVF International Conference on Computer Vision (ICCV)}}.
\newblock


\bibitem[Kim et~al\mbox{.}(2022)]%
        {kim2022infonerf}
\bibfield{author}{\bibinfo{person}{Mijeong Kim}, \bibinfo{person}{Seonguk Seo},
  {and} \bibinfo{person}{Bohyung Han}.} \bibinfo{year}{2022}\natexlab{}.
\newblock \showarticletitle{{InfoNeRF}: Ray Entropy Minimization for Few-Shot
  Neural Volume Rendering}. In \bibinfo{booktitle}{\emph{Proceedings of the
  IEEE/CVF Conference on Computer Vision and Pattern Recognition (CVPR)}}.
\newblock


\bibitem[Kwak et~al\mbox{.}(2023)]%
        {kwak2023geconerf}
\bibfield{author}{\bibinfo{person}{Minseop Kwak}, \bibinfo{person}{Jiuhn Song},
  {and} \bibinfo{person}{Seungryong Kim}.} \bibinfo{year}{2023}\natexlab{}.
\newblock \showarticletitle{{GeCoNeRF}: Few-shot Neural Radiance Fields via
  Geometric Consistency}.
\newblock \bibinfo{journal}{\emph{arXiv e-prints}}, Article
  \bibinfo{articleno}{arXiv:2301.10941} (\bibinfo{year}{2023}),
  \bibinfo{numpages}{arXiv:2301.10941}~pages.
\newblock
\showeprint[arxiv]{2301.10941}


\bibitem[Lee et~al\mbox{.}(2023)]%
        {lee2023extremenerf}
\bibfield{author}{\bibinfo{person}{SeokYeong Lee}, \bibinfo{person}{JunYong
  Choi}, \bibinfo{person}{Seungryong Kim}, \bibinfo{person}{Ig-Jae Kim}, {and}
  \bibinfo{person}{Junghyun Cho}.} \bibinfo{year}{2023}\natexlab{}.
\newblock \showarticletitle{{ExtremeNeRF}: Few-shot Neural Radiance Fields
  Under Unconstrained Illumination}.
\newblock \bibinfo{journal}{\emph{arXiv e-prints}}, Article
  \bibinfo{articleno}{arXiv:2303.11728} (\bibinfo{year}{2023}),
  \bibinfo{numpages}{arXiv:2303.11728}~pages.
\newblock
\showeprint[arxiv]{2303.11728}


\bibitem[Levoy and Hanrahan(1996)]%
        {levoy1996light}
\bibfield{author}{\bibinfo{person}{Marc Levoy} {and} \bibinfo{person}{Pat
  Hanrahan}.} \bibinfo{year}{1996}\natexlab{}.
\newblock \showarticletitle{Light Field Rendering}. In
  \bibinfo{booktitle}{\emph{Proceedings of the Computer Graphics and
  Interactive Techniques (SIGGRAPH)}}.
\newblock
\urldef\tempurl%
\url{https://doi.org/10.1145/237170.237199}
\showDOI{\tempurl}


\bibitem[Lin et~al\mbox{.}(2023)]%
        {lin2023vision}
\bibfield{author}{\bibinfo{person}{Kai-En Lin}, \bibinfo{person}{Yen-Chen Lin},
  \bibinfo{person}{Wei-Sheng Lai}, \bibinfo{person}{Tsung-Yi Lin},
  \bibinfo{person}{Yi-Chang Shih}, {and} \bibinfo{person}{Ravi Ramamoorthi}.}
  \bibinfo{year}{2023}\natexlab{}.
\newblock \showarticletitle{Vision Transformer for {NeRF}-Based View Synthesis
  From a Single Input Image}. In \bibinfo{booktitle}{\emph{Proceedings of the
  IEEE/CVF Winter Conference on Applications of Computer Vision (WACV)}}.
\newblock


\bibitem[Ma et~al\mbox{.}(2022)]%
        {ma2022deblurnerf}
\bibfield{author}{\bibinfo{person}{Li Ma}, \bibinfo{person}{Xiaoyu Li},
  \bibinfo{person}{Jing Liao}, \bibinfo{person}{Qi Zhang},
  \bibinfo{person}{Xuan Wang}, \bibinfo{person}{Jue Wang}, {and}
  \bibinfo{person}{Pedro~V. Sander}.} \bibinfo{year}{2022}\natexlab{}.
\newblock \showarticletitle{{Deblur-NeRF}: Neural Radiance Fields From Blurry
  Images}. In \bibinfo{booktitle}{\emph{Proceedings of the IEEE/CVF Conference
  on Computer Vision and Pattern Recognition (CVPR)}}.
\newblock


\bibitem[Menapace et~al\mbox{.}(2023)]%
        {menapace2023plotting}
\bibfield{author}{\bibinfo{person}{Willi Menapace}, \bibinfo{person}{Aliaksandr
  Siarohin}, \bibinfo{person}{St{\'e}phane Lathuili{\`e}re},
  \bibinfo{person}{Panos Achlioptas}, \bibinfo{person}{Vladislav Golyanik},
  \bibinfo{person}{Elisa Ricci}, {and} \bibinfo{person}{Sergey Tulyakov}.}
  \bibinfo{year}{2023}\natexlab{}.
\newblock \showarticletitle{Plotting Behind the Scenes: Towards Learnable Game
  Engines}.
\newblock \bibinfo{journal}{\emph{arXiv e-prints}}, Article
  \bibinfo{articleno}{arXiv:2303.13472} (\bibinfo{year}{2023}),
  \bibinfo{numpages}{arXiv:2303.13472}~pages.
\newblock
\showeprint[arxiv]{2303.13472}


\bibitem[Mildenhall et~al\mbox{.}(2022)]%
        {mildenhall2022nerfinthedark}
\bibfield{author}{\bibinfo{person}{Ben Mildenhall}, \bibinfo{person}{Peter
  Hedman}, \bibinfo{person}{Ricardo Martin-Brualla}, \bibinfo{person}{Pratul~P.
  Srinivasan}, {and} \bibinfo{person}{Jonathan~T. Barron}.}
  \bibinfo{year}{2022}\natexlab{}.
\newblock \showarticletitle{NeRF in the Dark: High Dynamic Range View Synthesis
  From Noisy Raw Images}. In \bibinfo{booktitle}{\emph{Proceedings of the
  IEEE/CVF Conference on Computer Vision and Pattern Recognition (CVPR)}}.
\newblock


\bibitem[Mildenhall et~al\mbox{.}(2019)]%
        {mildenhall2019llff}
\bibfield{author}{\bibinfo{person}{Ben Mildenhall}, \bibinfo{person}{Pratul~P.
  Srinivasan}, \bibinfo{person}{Rodrigo Ortiz-Cayon},
  \bibinfo{person}{Nima~Khademi Kalantari}, \bibinfo{person}{Ravi Ramamoorthi},
  \bibinfo{person}{Ren Ng}, {and} \bibinfo{person}{Abhishek Kar}.}
  \bibinfo{year}{2019}\natexlab{}.
\newblock \showarticletitle{Local Light Field Fusion: Practical View Synthesis
  with Prescriptive Sampling Guidelines}.
\newblock \bibinfo{journal}{\emph{ACM Transactions on Graphics (TOG)}}
  \bibinfo{volume}{38}, \bibinfo{number}{4} (\bibinfo{date}{July}
  \bibinfo{year}{2019}), \bibinfo{pages}{1--14}.
\newblock
\urldef\tempurl%
\url{https://doi.org/10.1145/3306346.3322980}
\showDOI{\tempurl}


\bibitem[Mildenhall et~al\mbox{.}(2020)]%
        {mildenhall2020nerf}
\bibfield{author}{\bibinfo{person}{Ben Mildenhall}, \bibinfo{person}{Pratul~P.
  Srinivasan}, \bibinfo{person}{Matthew Tancik}, \bibinfo{person}{Jonathan~T.
  Barron}, \bibinfo{person}{Ravi Ramamoorthi}, {and} \bibinfo{person}{Ren Ng}.}
  \bibinfo{year}{2020}\natexlab{}.
\newblock \showarticletitle{NeRF: Representing Scenes as Neural Radiance Fields
  for View Synthesis}. In \bibinfo{booktitle}{\emph{Proceedings of the European
  Conference on Computer Vision (ECCV)}}.
\newblock


\bibitem[M{\"u}ller et~al\mbox{.}(2022)]%
        {muller2022instant}
\bibfield{author}{\bibinfo{person}{Thomas M{\"u}ller}, \bibinfo{person}{Alex
  Evans}, \bibinfo{person}{Christoph Schied}, {and} \bibinfo{person}{Alexander
  Keller}.} \bibinfo{year}{2022}\natexlab{}.
\newblock \showarticletitle{Instant Neural Graphics Primitives with a
  Multiresolution Hash Encoding}.
\newblock \bibinfo{journal}{\emph{ACM Transactions on Graphics (ToG)}}
  \bibinfo{volume}{41}, \bibinfo{number}{4} (\bibinfo{year}{2022}),
  \bibinfo{pages}{1--15}.
\newblock


\bibitem[Niemeyer et~al\mbox{.}(2022)]%
        {niemeyer2022regnerf}
\bibfield{author}{\bibinfo{person}{Michael Niemeyer},
  \bibinfo{person}{Jonathan~T. Barron}, \bibinfo{person}{Ben Mildenhall},
  \bibinfo{person}{Mehdi S.~M. Sajjadi}, \bibinfo{person}{Andreas Geiger},
  {and} \bibinfo{person}{Noha Radwan}.} \bibinfo{year}{2022}\natexlab{}.
\newblock \showarticletitle{{RegNeRF}: Regularizing Neural Radiance Fields for
  View Synthesis From Sparse Inputs}. In \bibinfo{booktitle}{\emph{Proceedings
  of the IEEE/CVF Conference on Computer Vision and Pattern Recognition
  (CVPR)}}.
\newblock


\bibitem[Pearl et~al\mbox{.}(2022)]%
        {pearl2022nan}
\bibfield{author}{\bibinfo{person}{Naama Pearl}, \bibinfo{person}{Tali
  Treibitz}, {and} \bibinfo{person}{Simon Korman}.}
  \bibinfo{year}{2022}\natexlab{}.
\newblock \showarticletitle{{NAN}: Noise-Aware NeRFs for Burst-Denoising}. In
  \bibinfo{booktitle}{\emph{Proceedings of the IEEE/CVF Conference on Computer
  Vision and Pattern Recognition (CVPR)}}.
\newblock


\bibitem[Roessle et~al\mbox{.}(2022)]%
        {roessle2022ddpnerf}
\bibfield{author}{\bibinfo{person}{Barbara Roessle},
  \bibinfo{person}{Jonathan~T. Barron}, \bibinfo{person}{Ben Mildenhall},
  \bibinfo{person}{Pratul~P. Srinivasan}, {and} \bibinfo{person}{Matthias
  Nie{\ss}ner}.} \bibinfo{year}{2022}\natexlab{}.
\newblock \showarticletitle{Dense Depth Priors for Neural Radiance Fields From
  Sparse Input Views}. In \bibinfo{booktitle}{\emph{Proceedings of the IEEE/CVF
  Conference on Computer Vision and Pattern Recognition (CVPR)}}.
\newblock


\bibitem[Seo et~al\mbox{.}(2023)]%
        {seo2023mixnerf}
\bibfield{author}{\bibinfo{person}{Seunghyeon Seo}, \bibinfo{person}{Donghoon
  Han}, \bibinfo{person}{Yeonjin Chang}, {and} \bibinfo{person}{Nojun Kwak}.}
  \bibinfo{year}{2023}\natexlab{}.
\newblock \showarticletitle{{MixNeRF}: Modeling a Ray with Mixture Density for
  Novel View Synthesis from Sparse Inputs}. In
  \bibinfo{booktitle}{\emph{Proceedings of the IEEE/CVF Conference on Computer
  Vision and Pattern Recognition (CVPR)}}.
\newblock


\bibitem[Shi et~al\mbox{.}(2022)]%
        {shi2022garf}
\bibfield{author}{\bibinfo{person}{Yue Shi}, \bibinfo{person}{Dingyi Rong},
  \bibinfo{person}{Bingbing Ni}, \bibinfo{person}{Chang Chen}, {and}
  \bibinfo{person}{Wenjun Zhang}.} \bibinfo{year}{2022}\natexlab{}.
\newblock \showarticletitle{{GARF}: Geometry-Aware Generalized Neural Radiance
  Field}.
\newblock \bibinfo{journal}{\emph{arXiv e-prints}}, Article
  \bibinfo{articleno}{arXiv:2212.02280} (\bibinfo{year}{2022}),
  \bibinfo{numpages}{arXiv:2212.02280}~pages.
\newblock
\showeprint[arxiv]{2212.02280}


\bibitem[Somraj and Soundararajan(2023)]%
        {somraj2023vipnerf}
\bibfield{author}{\bibinfo{person}{Nagabhushan Somraj} {and}
  \bibinfo{person}{Rajiv Soundararajan}.} \bibinfo{year}{2023}\natexlab{}.
\newblock \showarticletitle{{ViP-NeRF}: Visibility Prior for Sparse Input
  Neural Radiance Fields}. In \bibinfo{booktitle}{\emph{Proceedings of the ACM
  Special Interest Group on Computer Graphics and Interactive Techniques
  (SIGGRAPH)}}.
\newblock
\urldef\tempurl%
\url{https://doi.org/10.1145/3588432.3591539}
\showDOI{\tempurl}


\bibitem[Tancik et~al\mbox{.}(2021)]%
        {tancik2021metanerf}
\bibfield{author}{\bibinfo{person}{Matthew Tancik}, \bibinfo{person}{Ben
  Mildenhall}, \bibinfo{person}{Terrance Wang}, \bibinfo{person}{Divi Schmidt},
  \bibinfo{person}{Pratul~P. Srinivasan}, \bibinfo{person}{Jonathan~T. Barron},
  {and} \bibinfo{person}{Ren Ng}.} \bibinfo{year}{2021}\natexlab{}.
\newblock \showarticletitle{Learned Initializations for Optimizing
  Coordinate-Based Neural Representations}. In
  \bibinfo{booktitle}{\emph{Proceedings of the IEEE/CVF Conference on Computer
  Vision and Pattern Recognition (CVPR)}}.
\newblock


\bibitem[Trevithick and Yang(2021)]%
        {trevithick2021grf}
\bibfield{author}{\bibinfo{person}{Alex Trevithick} {and} \bibinfo{person}{Bo
  Yang}.} \bibinfo{year}{2021}\natexlab{}.
\newblock \showarticletitle{{GRF}: Learning a General Radiance Field for {3D}
  Representation and Rendering}. In \bibinfo{booktitle}{\emph{Proceedings of
  the IEEE/CVF International Conference on Computer Vision (ICCV)}}.
\newblock


\bibitem[Uy et~al\mbox{.}(2023)]%
        {uy2023scade}
\bibfield{author}{\bibinfo{person}{Mikaela~Angelina Uy},
  \bibinfo{person}{Ricardo Martin-Brualla}, \bibinfo{person}{Leonidas Guibas},
  {and} \bibinfo{person}{Ke Li}.} \bibinfo{year}{2023}\natexlab{}.
\newblock \showarticletitle{SCADE: NeRFs from Space Carving with
  Ambiguity-Aware Depth Estimates}.
\newblock  (\bibinfo{date}{June} \bibinfo{year}{2023}).
\newblock


\bibitem[Verbin et~al\mbox{.}(2022)]%
        {verbin2022refnerf}
\bibfield{author}{\bibinfo{person}{Dor Verbin}, \bibinfo{person}{Peter Hedman},
  \bibinfo{person}{Ben Mildenhall}, \bibinfo{person}{Todd Zickler},
  \bibinfo{person}{Jonathan~T. Barron}, {and} \bibinfo{person}{Pratul~P.
  Srinivasan}.} \bibinfo{year}{2022}\natexlab{}.
\newblock \showarticletitle{{Ref-NeRF}: Structured View-Dependent Appearance
  for Neural Radiance Fields}. In \bibinfo{booktitle}{\emph{Proceedings of the
  IEEE/CVF Conference on Computer Vision and Pattern Recognition (CVPR)}}.
\newblock
\urldef\tempurl%
\url{https://doi.org/10.1109/CVPR52688.2022.00541}
\showDOI{\tempurl}


\bibitem[Wang et~al\mbox{.}(2023)]%
        {wang2023sparsenerf}
\bibfield{author}{\bibinfo{person}{Guangcong Wang}, \bibinfo{person}{Zhaoxi
  Chen}, \bibinfo{person}{Chen~Change Loy}, {and} \bibinfo{person}{Ziwei Liu}.}
  \bibinfo{year}{2023}\natexlab{}.
\newblock \showarticletitle{{SparseNeRF}: Distilling Depth Ranking for Few-shot
  Novel View Synthesis}.
\newblock \bibinfo{journal}{\emph{arXiv e-prints}}, Article
  \bibinfo{articleno}{arXiv:2303.16196} (\bibinfo{year}{2023}),
  \bibinfo{numpages}{arXiv:2303.16196}~pages.
\newblock
\showeprint[arxiv]{2303.16196}


\bibitem[Wang et~al\mbox{.}(2021)]%
        {wang2021ibrnet}
\bibfield{author}{\bibinfo{person}{Qianqian Wang}, \bibinfo{person}{Zhicheng
  Wang}, \bibinfo{person}{Kyle Genova}, \bibinfo{person}{Pratul~P. Srinivasan},
  \bibinfo{person}{Howard Zhou}, \bibinfo{person}{Jonathan~T. Barron},
  \bibinfo{person}{Ricardo Martin-Brualla}, \bibinfo{person}{Noah Snavely},
  {and} \bibinfo{person}{Thomas Funkhouser}.} \bibinfo{year}{2021}\natexlab{}.
\newblock \showarticletitle{{IBRNet}: Learning Multi-View Image-Based
  Rendering}. In \bibinfo{booktitle}{\emph{Proceedings of the IEEE/CVF
  Conference on Computer Vision and Pattern Recognition (CVPR)}}.
\newblock


\bibitem[Wang et~al\mbox{.}(2004)]%
        {wang2004image}
\bibfield{author}{\bibinfo{person}{Zhou Wang}, \bibinfo{person}{Alan~C Bovik},
  \bibinfo{person}{Hamid~R Sheikh}, {and} \bibinfo{person}{Eero~P Simoncelli}.}
  \bibinfo{year}{2004}\natexlab{}.
\newblock \showarticletitle{Image quality assessment: from error visibility to
  structural similarity}.
\newblock \bibinfo{journal}{\emph{IEEE Transactions on Image Processing (TIP)}}
  \bibinfo{volume}{13}, \bibinfo{number}{4} (\bibinfo{year}{2004}),
  \bibinfo{pages}{600--612}.
\newblock


\bibitem[Wynn and Turmukhambetov(2023)]%
        {wynn2023diffusionerf}
\bibfield{author}{\bibinfo{person}{Jamie Wynn} {and} \bibinfo{person}{Daniyar
  Turmukhambetov}.} \bibinfo{year}{2023}\natexlab{}.
\newblock \showarticletitle{{DiffusioNeRF}: Regularizing Neural Radiance Fields
  with Denoising Diffusion Models}.
\newblock \bibinfo{journal}{\emph{arXiv e-prints}}, Article
  \bibinfo{articleno}{arXiv:2302.12231} (\bibinfo{year}{2023}),
  \bibinfo{numpages}{arXiv:2302.12231}~pages.
\newblock
\showeprint[arxiv]{2302.12231}


\bibitem[Xu et~al\mbox{.}(2022)]%
        {xu2022sinnerf}
\bibfield{author}{\bibinfo{person}{Dejia Xu}, \bibinfo{person}{Yifan Jiang},
  \bibinfo{person}{Peihao Wang}, \bibinfo{person}{Zhiwen Fan},
  \bibinfo{person}{Humphrey Shi}, {and} \bibinfo{person}{Zhangyang Wang}.}
  \bibinfo{year}{2022}\natexlab{}.
\newblock \showarticletitle{{SinNeRF}: Training Neural Radiance Fields on
  Complex Scenes from a Single Image}. In \bibinfo{booktitle}{\emph{Proceedings
  of the European Conference on Computer Vision (ECCV)}}.
\newblock


\bibitem[Yang et~al\mbox{.}(2023)]%
        {yang2023freenerf}
\bibfield{author}{\bibinfo{person}{Jiawei Yang}, \bibinfo{person}{Marco
  Pavone}, {and} \bibinfo{person}{Yue Wang}.} \bibinfo{year}{2023}\natexlab{}.
\newblock \showarticletitle{{FreeNeRF}: Improving Few-shot Neural Rendering
  with Free Frequency Regularization}.
\newblock  (\bibinfo{date}{June} \bibinfo{year}{2023}).
\newblock


\bibitem[Yariv et~al\mbox{.}(2021)]%
        {yariv2021volume}
\bibfield{author}{\bibinfo{person}{Lior Yariv}, \bibinfo{person}{Jiatao Gu},
  \bibinfo{person}{Yoni Kasten}, {and} \bibinfo{person}{Yaron Lipman}.}
  \bibinfo{year}{2021}\natexlab{}.
\newblock \showarticletitle{Volume Rendering of Neural Implicit Surfaces}. In
  \bibinfo{booktitle}{\emph{Proceedings of the Advances in Neural Information
  Processing Systems (NeurIPS)}}.
\newblock


\bibitem[Yu et~al\mbox{.}(2021)]%
        {yu2021pixelnerf}
\bibfield{author}{\bibinfo{person}{Alex Yu}, \bibinfo{person}{Vickie Ye},
  \bibinfo{person}{Matthew Tancik}, {and} \bibinfo{person}{Angjoo Kanazawa}.}
  \bibinfo{year}{2021}\natexlab{}.
\newblock \showarticletitle{{pixelNeRF}: Neural Radiance Fields From One or Few
  Images}. In \bibinfo{booktitle}{\emph{Proceedings of the IEEE Conference on
  Computer Vision and Pattern Recognition (CVPR)}}.
\newblock


\bibitem[Yuan et~al\mbox{.}(2022)]%
        {yuan2022nerfediting}
\bibfield{author}{\bibinfo{person}{Yu-Jie Yuan}, \bibinfo{person}{Yang-Tian
  Sun}, \bibinfo{person}{Yu-Kun Lai}, \bibinfo{person}{Yuewen Ma},
  \bibinfo{person}{Rongfei Jia}, {and} \bibinfo{person}{Lin Gao}.}
  \bibinfo{year}{2022}\natexlab{}.
\newblock \showarticletitle{NeRF-Editing: Geometry Editing of Neural Radiance
  Fields}. In \bibinfo{booktitle}{\emph{Proceedings of the IEEE/CVF Conference
  on Computer Vision and Pattern Recognition (CVPR)}}.
\newblock


\bibitem[Zhang et~al\mbox{.}(2021)]%
        {zhang2021ners}
\bibfield{author}{\bibinfo{person}{Jason Zhang}, \bibinfo{person}{Gengshan
  Yang}, \bibinfo{person}{Shubham Tulsiani}, {and} \bibinfo{person}{Deva
  Ramanan}.} \bibinfo{year}{2021}\natexlab{}.
\newblock \showarticletitle{{NeRS}: Neural Reflectance Surfaces for Sparse-view
  {3D} Reconstruction in the Wild}. In \bibinfo{booktitle}{\emph{Proceedings of
  the Advances in Neural Information Processing Systems (NeurIPS)}}.
\newblock


\bibitem[Zhang et~al\mbox{.}(2020)]%
        {zhang2020nerfpp}
\bibfield{author}{\bibinfo{person}{Kai Zhang}, \bibinfo{person}{Gernot
  Riegler}, \bibinfo{person}{Noah Snavely}, {and} \bibinfo{person}{Vladlen
  Koltun}.} \bibinfo{year}{2020}\natexlab{}.
\newblock \showarticletitle{{NeRF++}: Analyzing and Improving Neural Radiance
  Fields}.
\newblock \bibinfo{journal}{\emph{arXiv e-prints}}, Article
  \bibinfo{articleno}{arXiv:2010.07492} (\bibinfo{year}{2020}),
  \bibinfo{numpages}{arXiv:2010.07492}~pages.
\newblock
\showeprint[arxiv]{2010.07492}


\bibitem[Zhang et~al\mbox{.}(2018)]%
        {zhang2018unreasonable}
\bibfield{author}{\bibinfo{person}{Richard Zhang}, \bibinfo{person}{Phillip
  Isola}, \bibinfo{person}{Alexei~A Efros}, \bibinfo{person}{Eli Shechtman},
  {and} \bibinfo{person}{Oliver Wang}.} \bibinfo{year}{2018}\natexlab{}.
\newblock \showarticletitle{The Unreasonable Effectiveness of Deep Features as
  a Perceptual Metric}. In \bibinfo{booktitle}{\emph{Proceedings of the IEEE
  Conference on Computer Vision and Pattern Recognition (CVPR)}}.
\newblock


\bibitem[Zhou et~al\mbox{.}(2018)]%
        {zhou2018stereomag}
\bibfield{author}{\bibinfo{person}{Tinghui Zhou}, \bibinfo{person}{Richard
  Tucker}, \bibinfo{person}{John Flynn}, \bibinfo{person}{Graham Fyffe}, {and}
  \bibinfo{person}{Noah Snavely}.} \bibinfo{year}{2018}\natexlab{}.
\newblock \showarticletitle{Stereo Magnification: Learning View Synthesis Using
  Multiplane Images}.
\newblock \bibinfo{journal}{\emph{ACM Transactions on Graphics (TOG)}}
  \bibinfo{volume}{37}, \bibinfo{number}{4} (\bibinfo{date}{July}
  \bibinfo{year}{2018}).
\newblock


\bibitem[Zhu et~al\mbox{.}(2023)]%
        {zhu2023vdnnerf}
\bibfield{author}{\bibinfo{person}{Bingfan Zhu}, \bibinfo{person}{Yanchao
  Yang}, \bibinfo{person}{Xulong Wang}, \bibinfo{person}{Youyi Zheng}, {and}
  \bibinfo{person}{Leonidas Guibas}.} \bibinfo{year}{2023}\natexlab{}.
\newblock \showarticletitle{{VDN-NeRF}: Resolving Shape-Radiance Ambiguity via
  View-Dependence Normalization}.
\newblock \bibinfo{journal}{\emph{arXiv e-prints}}, Article
  \bibinfo{articleno}{arXiv:2303.17968} (\bibinfo{year}{2023}),
  \bibinfo{numpages}{arXiv:2303.17968}~pages.
\newblock
\showeprint[arxiv]{2303.17968}


\end{thebibliography}

\appendix
\twocolumn[\subsection*{\centering \fontsize{13}{15}\selectfont Supplement}]

    \noindent The contents of this supplement include
    \begin{enumerate}[label=\Alph*., noitemsep]
        \item Details on databases, evaluation, and implementations.
        \item Video examples on LLFF and RealEstate-10K datasets.
        \item Discussions on FreeNeRF, RegNeRF and ViP-NeRF\@.
        \item Additional comparisons.
        \item Additional analysis - positional encoding frequency, depth reliability masks and comparison with vanilla NeRF\@.
        \item Extensive quantitative evaluation reports.
    \end{enumerate}

    \section{Database, Evaluation, and Implementation Details}\label{sec:details-database-evaluation-implementation}
    \subsection{Database Details}\label{subsec:database-details}
    \begin{table}
        \centering
        \caption{Train and test frame numbers of RealEstate-10K dataset used in the three different settings.}
        \begin{tabular}{|c|c|c|}
            \hline
            No. of i/p frames & Train frame nos.\ & Test frame nos.\ \\
            \hline
            2 & 10, 20        & 5--9, 11--19, 21--25 \\
            3 & 10, 20, 30    & 5--9, 11--19, 21--29, 31--35 \\
            4 & 0, 10, 20, 30 & 1--9, 11--19, 21--29, 31--35 \\
            \hline
        \end{tabular}
        \label{tab:realestate-train-test}
    \end{table}
    We compare the performance of sparse input NeRF models on LLFF and RealEstate-10K datasets with $n = 2, 3, 4$ input views.

    LLFF~\cite{mildenhall2019llff} dataset contains 8 forward-facing unbounded scenes.
    Each scene contains a variable number of frames at a spatial resolution of $1008 \times 756$.
    Following prior art~\cite{niemeyer2022regnerf}, we use every $8^\text{th}$ frame for testing in each scene.
    We uniformly sample $n$ training views from the remaining frames.

    RealEstate-10K~\cite{zhou2018stereomag} is a dataset of videos with camera motion popularly used for novel view synthesis.
    From its large test set, we choose five scenes following ViP-NeRF~\cite{somraj2023vipnerf}.
    Each scene contains 50 temporally continuous frames at a spatial resolution of $1024 \times 576$.
    We reserve frames 10, 20, 30, 0, and 40 for training and the remaining 45 frames for testing in every scene.
    We use the first $n$ frames from the above list as training views for $n$-view NeRF\@.
    For testing, we choose all the frames between the training views that correspond to interpolation and five frames on either side that correspond to extrapolation.
    \cref{tab:realestate-train-test} provides exact details on train and test frames.


    \subsection{Evaluation Details}\label{subsec:evaluation-details}
    As mentioned in the main paper, we evaluate the model predictions only in the regions visible in the training images.
    We now explain our reasoning behind masking the model predictions for evaluation and then provide the details of how we compute the masks.

    Recall that NeRFs are designed to memorize a scene and are therefore not equipped to predict unseen regions by design.
    Further, many regularization based sparse input NeRF models do not employ pre-trained prior.
    As a result, NeRFs are also ill-equipped to predict the depth of objects seen in only one of the input views, again by design.
    Thus, NeRFs require the objects to be visible in at least two views to estimate their 3D geometry accurately.
    Hence, we generate a mask that denotes the pixels visible in at least two input views, and evaluate the NeRF predictions in such regions only.

    We generate the mask by using the depths predicted by the dense input NeRF model as follows.
    For every train view, we warp its depth predicted by Dense-NeRF to every other test view and compare it with the Dense-NeRF predicted depth of the test view.
    Intuitively, if a pixel in a test view is visible in the considered train view, then the two depths should be close to each other.
    Thus, we threshold the depth difference to obtain the mask.
    That is, if the difference in the two depths is less than a threshold, then we mark the pixel in the test view as visible in the considered train view.
    Our final mask for every test view is generated by marking pixels as visible if they are visible in at least two input views.
    We warp the depth maps using depth based reprojection similar to \citet{cho2017hole,kanchana2022ivp} and set the threshold to $0.05$ times the maximum depth of the train view.
    By computing the mask using depths and not color, our approach is robust to the presence of specular objects, as long as the depth estimated by Dense-NeRF is accurate.
    We will release the code used to generate the masks and the masks along with the main code release.

    Nonetheless, we report the performance without masking the unseen regions in \cref{tab:quantitative-all-llff01,tab:quantitative-all-llff02,tab:quantitative-all-llff03,tab:quantitative-all-realestate01,tab:quantitative-all-realestate02,tab:quantitative-all-realestate03}.

    \subsection{Implementation Details}\label{subsec:implementation-details}
    We train the competing models using the official code releases by the respective authors using the configurations recommended in the code releases.
    Our code is developed in PyTorch and on top of DS-NeRF~\cite{deng2022dsnerf}.
    We employ Adam Optimizer with an initial learning rate of 5e-4 and exponentially decay it to 5e-6.
    We adjust the weights for the different losses such that their magnitudes after scaling are of a similar order.
    For the first 10k iterations of the training, we impose $\losscolor$ and $\losssd$ only.
    $\lossap, \lossav$ and $\losscfc$ are imposed after 10k iterations.
    We set the hyper-parameters as follows:
    $l_p = 10, l_v = 4, l_p^{\text{ap}} = 3, k=5, e_\tau = 0.1, \lambda_1 = 1, \text{ and } \lambda_2 = \lambda_3 = \lambda_4 = \lambda_5 = 0.1$.
    The network architecture is exactly the same as DS-NeRF\@.
    For the augmented models, we only change the input dimension of the MLPs $\mlpf_1$ and $\mlpf_2$ appropriately.
    The augmented models are employed only during training, and the network is exactly the same as Vanilla NeRF for inference.
    We train the models on a single NVIDIA RTX A4000 16GB GPU for 100k iterations.
    Our model takes about 22 hours to train per scene.

    \section{Video Comparisons}\label{sec:video-comparisons}
    We compare various models by rendering videos along a continuous trajectory.
    For the LLFF dataset, we render the videos along the spiral trajectory that is commonly used in the literature.
    Since RealEstate-10K is a dataset of videos, we combine the train and test frames to get the continuous trajectories.

    We divide the video comparisons into two sets.
    In the first set, we show how our regularizations reduce the floaters and shape-radiance ambiguity by comparing the videos rendered by our model with those of the competing models and the ablated models.
    In the second set, we compare the videos rendered by our model with those of the competing models.
    The videos are available on our project website \url{https://nagabhushansn95.github.io/publications/2023/SimpleNeRF.html}

    \section{Discussions on Competing Models}\label{sec:discussion-competing-models}
    \begin{figure*}
        \centering
        \includegraphics[width=\linewidth]{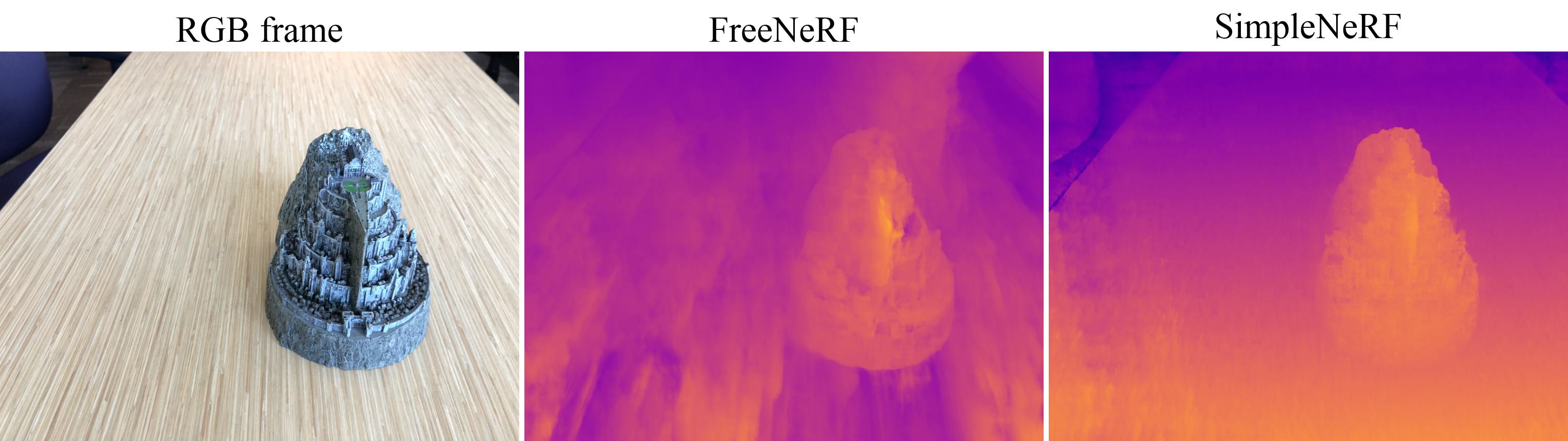}
        \caption{Comparison of depth maps predicted by FreeNeRF and SimpleNeRF on fortress scene of the LLFF dataset.
        Notice the distortions in the depth map estimated by FreeNeRF in the regions near the camera.
        SimpleNeRF does not suffer from such issues.
        }
        \label{fig:free-nerf-limitation}
    \end{figure*}
    \subsection{FreeNeRF}\label{subsec:freenerf}
    FreeNeRF~\cite{yang2023freenerf} is a concurrent work to ours and is also designed to train NeRFs with sparse input views.
    FreeNeRF also employs the idea of using fewer positional encoding frequencies.
    However, we developed our model independently, and is also sufficiently different from FreeNeRF\@.
    Firstly, our idea is based on using simpler solutions and is more general than FreeNeRF\@.
    For example, we employ view-independent color predicting NeRF as an augmentation, unlike FreeNeRF, which uses fewer positional encoding frequencies with respect to the viewing direction too.
    Our approach is a stronger regularization than reducing positional encoding frequencies and is also applicable in models that may not use positional encoding~\cite{chen2022tensorf,muller2022instant}.
    Secondly, FreeNeRF starts with a lower positional encoding degree and anneals it to a higher value as the training progresses.
    In contrast, we employ the simpler NeRF models (with fewer positional encoding frequencies or with view-independent radiance) as an augmentation and use it to obtain reliable depth supervision.
    Further, we employ reprojection error to filter out incorrect depth estimations by the augmented models and thus supervise using only the reliable depth predictions.
    As a result, in the later training stages of FreeNeRF, there is nothing constraining the model to not use the high frequency components.
    Our model does not suffer from this issue.
    Our points augmentation model is richer compared to FreeNeRF, since we include the residual positional encoded values to predict the color.
    This is highly useful in textured regions which are smooth in terms of depth, but contain sharp discontinuities in color as a function of position.
    Finally, we also include the coarse-fine consistency loss, $\losscfc$, which is not employed in FreeNeRF\@.

    We also note that FreeNeRF uses an additional regularization term to penalize the NeRF if it predicts any objects close to the camera.
    We believe this is a dataset bias and may not work in general.
    For example, many scenes from the LLFF dataset satisfy the above constraint, but this is not satisfied on the RealEstate-10K dataset.
    This may also be contributing to the poor performance of FreeNeRF on the RealEstate-10K dataset.
    An exception to the above assumption in the LLFF dataset is the fortress scene, which contains a table that extends toward the camera.
    \cref{fig:free-nerf-limitation} shows the depth map predicted by FreeNeRF for the fortress scene, which is unsurprisingly incorrect near the camera.
    This is further validated by the poor performance of FreeNeRF on the fortress scene as witnessed in \cref{tab:quantitative-scene-wise-llff02}.
    SimpleNeRF achieves a SSIM of 0.84 while FreeNeRF only achieves 0.62, with three input views on the fortress scene.

    \subsection{RegNeRF}\label{subsec:regnerf}
    RegNeRF~\cite{niemeyer2022regnerf} is a popular few-shot NeRF model that regularizes NeRF by imposing depth smoothness in the input and various hallucinated viewpoints.
    Our approach of using fewer positional encoding frequencies in the points augmentation model also promotes depth smoothness.
    However, the depth smoothness imposed by RegNeRF is uniformly applied across all pixels which probably tries to place all the objects in a single plane.
    The depth smoothness promoted by SimpleNeRF is different in terms of two aspects.
    Firstly, we use another network to impose smoothness which can still learn to place the objects at different depths and learn the desirable depth discontinuities.
    Secondly, we determine whether the depth estimated by the augmented model is accurate before using it to supervise the main NeRF model.
    These two reasons perhaps contribute for the superior performance of SimpleNeRF over RegNeRF\@.


    \subsection{ViP-NeRF}\label{subsec:vipnerf}
    We note that the quantitative results in Tabs.\ 1 and 2 in the main paper differ from the values reported in ViP-NeRF~\cite{somraj2023vipnerf}.
    In the following, we describe the reasons for this difference.
    \begin{enumerate}
        \item In ViP-NeRF, the quality evaluation metrics are computed on full frames.
        However, we exclude the regions not seen in the input views.
        We explain the reasoning behind this in the main paper Sec.\ 5 and the details in \cref{subsec:evaluation-details}.
        \item On the RealEstate-10K dataset, while we use the same train set as that of ViP-NeRF, we modify the test set as shown in \cref{tab:realestate-train-test}.
        We explain the reasoning behind reducing the test set in \cref{subsec:database-details}.
        More specifically, the test views that are very far away from the train views may contain large unobserved regions.
        Since the NeRF is not trained to reproduce such large unobserved regions, we exclude such extreme viewpoints.
    \end{enumerate}

    \section{Additional Comparisons}\label{sec:additional-comparisons}
    We compare with two other sparse-input NeRF models, namely, DietNeRF~\cite{jain2021dietnerf} and InfoNeRF~\cite{kim2022infonerf}.
    We also evaluate the performance of SimpleNeRF without using the residual high frequency positional encodings $\gamma(\pointp_i, l_p^{\text{ap}}, l_p)$ while predicting the color in the points augmentation model.
    In \cref{tab:quantitative-all-llff01,tab:quantitative-all-llff02,tab:quantitative-all-llff03,tab:quantitative-all-realestate01,tab:quantitative-all-realestate02,tab:quantitative-all-realestate03}, we report the performance of all the comparisons on both datasets with 2, 3, and 4 input views in terms of all the five evaluation measures.
    The values in parenthesis show the unmasked evaluation scores.

    \section{Additional Analysis}\label{sec:additional-analysis}
    \subsection{Positional Encoding Frequency}\label{subsec:positional-encoding-frequency}
    \begin{figure}
        \includegraphics[width=\linewidth]{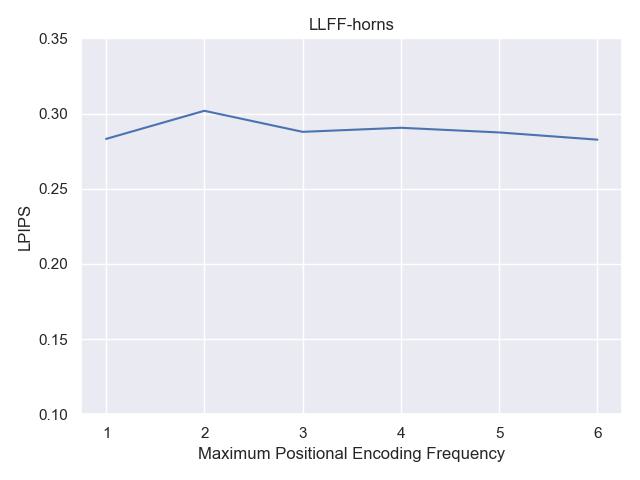}
        \caption{LPIPS scores on the horns scene as $l_p^{\text{ap}}$ is varied.}
        \label{fig:quantitative-positional-encoding-degree}
    \end{figure}
    We analyze the impact of the highest positional encoding frequency $l_p^{\text{ap}}$, used in the points augmentation.
    We vary $l_p^{\text{ap}}$ from $1$ to $6$ and test the performance of SimpleNeRF on the horns scene of the LLFF dataset.
    We show the quantitative performance in terms of LPIPS in \cref{fig:quantitative-positional-encoding-degree}.
    We observe only small variations in the performance as $l_p^{\text{ap}}$ is varied.
    Thus, our framework is robust to the choice of $l_p^{\text{ap}}$.
    We note that using $l_p^{\text{ap}} = l_p = 10$ is equivalent to using an identical augmentation.

    \subsection{Depth Reliability Masks Visualization}\label{subsec:depth-reliability-masks-visualization}
    \begin{figure*}
        \centering
        \begin{subfigure}[t]{0.48\linewidth}
            \centering
            \includegraphics[width=\linewidth]{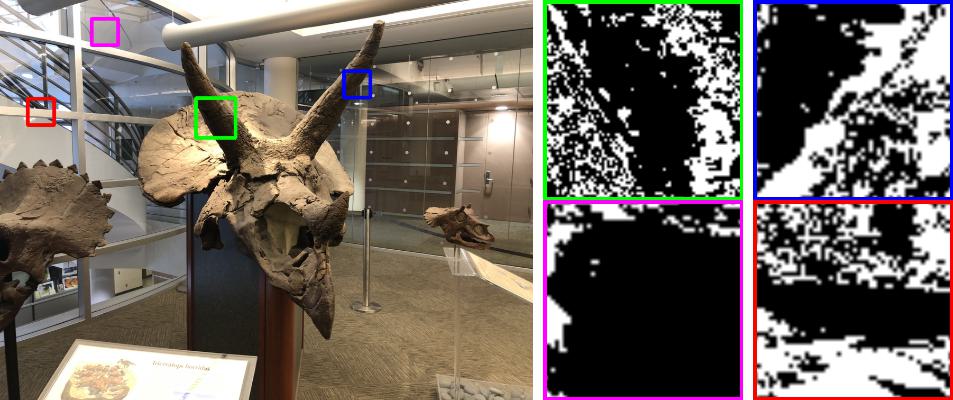}
            \caption{\textbf{Points augmentation:} The green and blue boxes focus on the two horns, where we observe that the augmented model depth is preferred in the depth-wise smooth regions on horns, and the main model depth is preferred at the edges. The magenta box focuses on a completely smooth region, so the augmented model depth is preferred for most pixels. In the red box, augmented model depth is preferred along the horizontal bar. The main model depth is preferred on either side of the bar that contains multiple depth discontinuities.}
            \label{fig:points-augmentation-visualization}
        \end{subfigure}
        \hfill
        \begin{subfigure}[t]{0.48\linewidth}
            \centering
            \includegraphics[width=\linewidth]{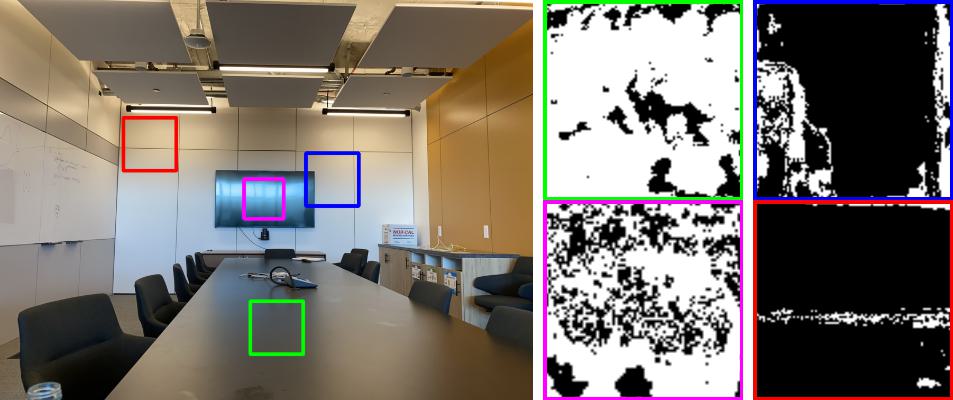}
            \caption{\textbf{Views augmentation:} The green and magenta boxes focus on the TV and the table, respectively, which are highly specular in this scene (please view the supplementary videos of the room scene to observe the specularity of these objects). In these regions, the main model depth is determined to be more accurate since the main model can handle specular regions. The red and blue boxes focus on Lambertian regions of the scene where the depth estimated by the augmented model is preferred.}
            \label{fig:views-augmentation-visualization}
        \end{subfigure}
        \caption{Visualizations of depth reliability mask for the two augmentations. White pixels in the mask indicate that the main model depth is determined to be more accurate at the corresponding locations. Black pixels indicate that the augmented model depth is determined to be more accurate.}
        \label{fig:augmentations-visualizations}
    \end{figure*}
    In \cref{fig:augmentations-visualizations}, we present visualizations that motivate the design of our augmentations, namely the points and views augmentations.
    We train our model without augmentations and the individual augmentations separately with only $\mathcal{L}_{\text{color}}$ and $\mathcal{L}_{\text{sd}}$ for 100k iterations.
    Using the depth maps predicted by the models for an input training view, we determine the mask that indicates which depth estimates are more accurate, as explained in Sec.\ 4.2.
    For two scenes from the LLFF dataset, we show an input training view and focus on a small region to visualize the corresponding masks.

    We observe that the points augmented model is determined to have estimated better depths in smooth regions.
    At edges, the depth estimated by the main model is more accurate.
    Similarly, the views augmented model estimates better depth in Lambertian regions, and the main model estimates better depth in specular regions.
    We note that the masks shown are not the masks obtained by our final model.
    Since the masks are computed at every iteration, and the training of the main and augmented models are coupled, it is not possible to determine the exact locations where the augmented models help the main model learn better.

    \subsection{Comparison with Vanilla NeRF}\label{subsec:vanilla-nerf}
    We analyze the impact of our regularizations when more input views are available.
    Specifically, we trained SimpleNeRF with all the available input views on the room scene of the LLFF dataset.
    We observed that the performance dropped a little lower than the Vanilla NeRF trained with the same input views, perhaps due to the following reason.
    During the initial part of the training, the augmented models may be more accurate and hence may influence the learning of the main model.
    As the training progresses, the augmented models do not learn the scene accurately due to their respective constraints.
    Since the augmented models constrain the main model in terms of estimated depths, it may lead to constraining the main model from freely learning the scene.
    This is perhaps the reason for a drop in the performance of SimpleNeRF as compared to the Vanilla NeRF model.

    \section{Performance on Individual Scenes}\label{sec:performance-on-individual-scenes}
    For the benefit of follow-up work, where researchers may want to analyze the performance of different models or compare the models on individual scenes, we provide the performance of various models on individual scenes in \cref{tab:quantitative-scene-wise-llff01,tab:quantitative-scene-wise-llff02,tab:quantitative-scene-wise-llff03,tab:quantitative-scene-wise-llff04,tab:quantitative-scene-wise-realestate01,tab:quantitative-scene-wise-realestate02,tab:quantitative-scene-wise-realestate03,tab:quantitative-scene-wise-realestate04}.

    \clearpage
    \begin{table*}
        \centering
        \caption{Quantitative Results on LLFF dataset with two input views. The values within parenthesis show unmasked scores.}
\begin{tabular}{l|ccc|cc}
            \hline
            \textbf{Model} & \textbf{LPIPS} & \textbf{SSIM} & \textbf{PSNR} & \textbf{Depth MAE}  & \textbf{Depth SROCC} \\
            \hline
            InfoNeRF     &  0.6024(0.7561) &  0.2219(0.2095) &  9.16(9.23) &  0.9797(1.1000) &  -0.0188(-0.0092)  \\
            DietNeRF     &  0.5465(0.7265) &  0.3283(0.3209) &  11.94(11.89) &  0.8886(1.0105) &  -0.0099(-0.0045)  \\
            RegNeRF      &  0.3056(0.4297) &  0.5712(0.4885) &  18.52(16.88) &  -- &  0.7141(0.6513)  \\
            DS-NeRF      &  0.3106(0.4176) &  0.5862(0.5074) &  18.24(16.93) &  0.2074(0.3372) &  0.7230(0.5787)  \\
            DDP-NeRF     &  0.2851(0.3920) &  0.6218(0.5424) &  18.73(17.19) &  0.2048(0.3494) &  0.7480(0.6109)  \\
            FreeNeRF     &  \textbf{0.2638}(0.3760) &  0.6322(0.5432) &  19.52(17.55) &  -- &  0.8066(0.7137)  \\
            ViP-NeRF     &  0.2768(\textbf{0.3725}) &  0.6225(0.5230) &  18.61(16.66) &  0.1999(0.3413) &  0.7344(0.6221)  \\
\hline
        \end{tabular}
        \label{tab:quantitative-all-llff01}
    \end{table*}

    \begin{table*}
        \centering
        \caption{Quantitative Results on LLFF dataset with three input views. The values within parenthesis show unmasked scores.}
        \begin{tabular}{l|ccc|cc}
            \hline
            \textbf{Model} & \textbf{LPIPS} & \textbf{SSIM} & \textbf{PSNR} & \textbf{Depth MAE}  & \textbf{Depth SROCC} \\
            \hline
            InfoNeRF     &  0.6732(0.7679) &  0.1953(0.1859) &  8.38(8.52) &  1.0012(1.1149) &  -0.0144(-0.0176)  \\
            DietNeRF     &  0.6120(0.7254) &  0.3405(0.3297) &  11.76(11.77) &  0.9093(1.0242) &  -0.0598(-0.0471)  \\
            RegNeRF      &  0.2908(0.3602) &  0.6334(0.5677) &  20.22(18.65) &  -- &  0.8238(0.7589)  \\
            DS-NeRF      &  0.3031(0.3641) &  0.6321(0.5774) &  20.20(18.97) &  0.1787(0.2699) &  0.7852(0.7173)  \\
            DDP-NeRF     &  0.3250(0.3869) &  0.6152(0.5628) &  18.73(17.71) &  0.1941(0.3032) &  0.7433(0.6707)  \\
            FreeNeRF     &  0.2754(0.3415) &  0.6583(0.5960) &  20.93(19.30) &  -- &  0.8379(0.7656)  \\
            ViP-NeRF     &  0.2798(0.3365) &  0.6548(0.5907) &  20.54(18.89) &  0.1721(0.2795) &  0.7891(0.7082)  \\
            \hline
            SimpleNeRF   &  \textbf{0.2559}(\textbf{0.3259}) &  \textbf{0.6940}(\textbf{0.6222}) &  \textbf{21.37}(\textbf{19.47}) &  \textbf{0.1199}(\textbf{0.2201}) &  \textbf{0.8935}(\textbf{0.8153})  \\
            \hline
        \end{tabular}
        \label{tab:quantitative-all-llff02}
    \end{table*}

    \begin{table*}
        \centering
        \caption{Quantitative Results on LLFF dataset with four input views. The values within parenthesis show unmasked scores.}
        \begin{tabular}{l|ccc|cc}
            \hline
            \textbf{Model} & \textbf{LPIPS} & \textbf{SSIM} & \textbf{PSNR} & \textbf{Depth MAE}  & \textbf{Depth SROCC} \\
            \hline
            InfoNeRF     &  0.6985(0.7701) &  0.2270(0.2188) &  9.18(9.25) &  1.0411(1.1119) &  -0.0394(-0.0390)  \\
            DietNeRF     &  0.6506(0.7396) &  0.3496(0.3404) &  11.86(11.84) &  0.9546(1.0259) &  -0.0368(-0.0249)  \\
            RegNeRF      &  0.2794(0.3227) &  0.6645(0.6159) &  21.32(19.89) &  -- &  0.8933(0.8528)  \\
            DS-NeRF      &  0.2979(0.3376) &  0.6582(0.6135) &  21.23(20.07) &  0.1451(0.2097) &  0.8506(0.8130)  \\
            DDP-NeRF     &  0.3042(0.3467) &  0.6558(0.6121) &  20.17(19.19) &  0.1704(0.2487) &  0.8322(0.7664)  \\
            FreeNeRF     &  0.2848(0.3280) &  0.6764(0.6303) &  21.91(\textbf{20.45}) &  -- &  0.9091(0.8626)  \\
            ViP-NeRF     &  0.2854(0.3203) &  0.6675(0.6182) &  20.75(19.34) &  0.1555(0.2316) &  0.8622(0.8070)  \\
            \hline
            SimpleNeRF   &  \textbf{0.2633}(\textbf{0.3083}) &  \textbf{0.7016}(\textbf{0.6521}) &  \textbf{21.99}(20.44) &  \textbf{0.1110}(\textbf{0.1741}) &  \textbf{0.9355}(\textbf{0.8952})  \\
            \hline
        \end{tabular}
        \label{tab:quantitative-all-llff03}
    \end{table*}

    \begin{table*}
        \centering
        \caption{Quantitative Results on RealEstate-10K dataset with two input views. The values within parenthesis show unmasked scores.}
        \begin{tabular}{l|ccc|cc}
            \hline
            \textbf{Model} & \textbf{LPIPS} & \textbf{SSIM} & \textbf{PSNR} & \textbf{Depth MAE}  & \textbf{Depth SROCC} \\
            \hline
            InfoNeRF     &  0.5924(0.6384) &  0.4342(0.4343) &  12.27(12.17) &  2.1793(2.2314) &  0.0912(0.0942)  \\
            DietNeRF     &  0.4381(0.4862) &  0.6534(0.6520) &  18.06(17.83) &  2.0247(2.0807) &  0.2339(0.2452)  \\
            RegNeRF      &  0.4129(0.4483) &  0.5916(0.5864) &  17.14(16.87) &  -- &  0.1118(0.1117)  \\
            DS-NeRF      &  0.2709(0.3171) &  0.7983(0.7859) &  26.26(25.44) &  0.7164(0.8323) &  0.6660(0.6433)  \\
            DDP-NeRF     &  0.1290(0.1481) &  0.8640(0.8502) &  27.79(26.15) &  0.4831(0.5944) &  0.7921(0.7663)  \\
            FreeNeRF     &  0.5036(0.5471) &  0.5354(0.5336) &  14.70(14.50) &  -- &  -0.1937(-0.1813)  \\
            ViP-NeRF     &  0.0687(0.0783) &  0.8889(0.8717) &  32.32(29.55) &  0.3856(0.5337) &  0.8446(0.7851)  \\
\hline
        \end{tabular}
        \label{tab:quantitative-all-realestate01}
    \end{table*}

    \begin{table*}
        \centering
        \caption{Quantitative Results on RealEstate-10K dataset with three input views. The values within parenthesis show unmasked scores.}
        \begin{tabular}{l|ccc|cc}
            \hline
            \textbf{Model} & \textbf{LPIPS} & \textbf{SSIM} & \textbf{PSNR} & \textbf{Depth MAE}  & \textbf{Depth SROCC} \\
            \hline
            InfoNeRF     &  0.6561(0.6846) &  0.3792(0.3780) &  10.57(10.57) &  2.2198(2.2830) &  0.1929(0.1994)  \\
            DietNeRF     &  0.4636(0.4886) &  0.6456(0.6445) &  18.01(17.89) &  2.0355(2.1023) &  0.0240(0.0438)  \\
            RegNeRF      &  0.4171(0.4362) &  0.6132(0.6078) &  17.86(17.73) &  -- &  0.0574(0.0475)  \\
            DS-NeRF      &  0.2893(0.3211) &  0.8004(0.7905) &  26.50(25.94) &  0.5400(0.6524) &  0.8106(0.7910)  \\
            DDP-NeRF     &  0.1518(0.1601) &  0.8587(0.8518) &  26.67(25.92) &  0.4139(0.5222) &  0.8612(0.8331)  \\
            FreeNeRF     &  0.5146(0.5414) &  0.5708(0.5675) &  15.26(15.12) &  -- &  -0.2590(-0.2445)  \\
            ViP-NeRF     &  0.0758(0.0832) &  0.8967(0.8852) &  31.93(30.27) &  0.3365(0.4683) &  0.9009(0.8558)  \\
            \hline
            SimpleNeRF   &  \textbf{0.0726}(\textbf{0.0829}) &  \textbf{0.8984}(\textbf{0.8879}) &  \textbf{33.21}(\textbf{31.40}) &  \textbf{0.2770}(\textbf{0.3885}) &  \textbf{0.9266}(\textbf{0.8931})  \\
            \hline
        \end{tabular}
        \label{tab:quantitative-all-realestate02}
    \end{table*}

    \begin{table*}
        \centering
        \caption{Quantitative Results on RealEstate-10K dataset with four input views. The values within parenthesis show unmasked scores.}
        \begin{tabular}{l|ccc|cc}
            \hline
            \textbf{Model} & \textbf{LPIPS} & \textbf{SSIM} & \textbf{PSNR} & \textbf{Depth MAE}  & \textbf{Depth SROCC} \\
            \hline
            InfoNeRF     &  0.6651(0.6721) &  0.3843(0.3830) &  10.62(10.59) &  2.1874(2.2742) &  0.2549(0.2594)  \\
            DietNeRF     &  0.4853(0.4954) &  0.6503(0.6475) &  18.01(17.89) &  2.0398(2.1273) &  0.0990(0.1011)  \\
            RegNeRF      &  0.4316(0.4383) &  0.6257(0.6198) &  18.34(18.25) &  -- &  0.1422(0.1396)  \\
            DS-NeRF      &  0.3103(0.3287) &  0.7999(0.7920) &  26.65(26.28) &  0.5154(0.6171) &  0.8145(0.8018)  \\
            DDP-NeRF     &  0.1563(0.1584) &  0.8617(0.8557) &  27.07(26.48) &  0.3832(0.4813) &  0.8739(0.8605)  \\
            FreeNeRF     &  0.5226(0.5323) &  0.6027(0.5989) &  16.31(16.25) &  -- &  -0.2152(-0.2162)  \\
            ViP-NeRF     &  0.0892(0.0909) &  0.8968(0.8894) &  31.95(30.83) &  0.3658(0.4761) &  0.8414(0.8080)  \\
            \hline
            SimpleNeRF   &  \textbf{0.0847}(\textbf{0.0891}) &  \textbf{0.8987}(\textbf{0.8917}) &  \textbf{32.88}(\textbf{31.73}) &  \textbf{0.2692}(\textbf{0.3565}) &  \textbf{0.9209}(\textbf{0.9035})  \\
            \hline
        \end{tabular}
        \label{tab:quantitative-all-realestate03}
    \end{table*}

    \begin{table*}
        \centering
        \caption{Per-scene performance of various models with two input views on LLFF dataset.
        The five rows show LPIPS, SSIM, PSNR, Depth MAE and Depth SROCC scores, respectively.
        The values within parenthesis show unmasked scores.}
        \begin{tabular}{l|c|c|c|c|c|c|c|c|c}
            \hline
            \textbf{\makecell{Model \textbackslash \\ Scene Name}} & \textbf{Fern} & \textbf{Flower} & \textbf{Fortress} & \textbf{Horns} & \textbf{Leaves} & \textbf{Orchids} & \textbf{Room} & \textbf{Trex} & \textbf{Average} \\
            \hline
            InfoNeRF     & \makecell{ 0.66(0.78)\\ 0.26(0.26)\\ 10.7(11.0)\\ 0.96(1.05)\\ -0.00(-0.04)} & \makecell{ 0.57(0.69)\\ 0.21(0.19)\\ 10.8(11.0)\\ 0.76(0.89)\\ -0.25(-0.18)} & \makecell{ 0.65(0.82)\\ 0.18(0.18)\\ 6.6(6.5)\\ 1.18(1.41)\\ 0.09(0.06)} & \makecell{ 0.58(0.76)\\ 0.20(0.20)\\ 8.9(8.9)\\ 1.08(1.26)\\ 0.07(0.08)} & \makecell{ 0.45(0.64)\\ 0.15(0.10)\\ 9.7(9.5)\\ 0.85(0.92)\\ -0.14(-0.12)} & \makecell{ 0.60(0.69)\\ 0.17(0.11)\\ 9.1(9.4)\\ 0.87(1.12)\\ 0.00(0.01)} & \makecell{ 0.68(0.80)\\ 0.36(0.38)\\ 10.4(10.8)\\ 1.05(1.06)\\ -0.01(0.00)} & \makecell{ 0.61(0.79)\\ 0.23(0.21)\\ 8.5(8.4)\\ 0.94(0.95)\\ -0.00(0.01)} & \makecell{ 0.60(0.76)\\ 0.22(0.21)\\ 9.2(9.2)\\ 0.98(1.10)\\ -0.02(-0.01)} \\
            \hline
            DietNeRF     & \makecell{ 0.63(0.77)\\ 0.30(0.29)\\ 12.1(12.3)\\ 0.83(0.92)\\ -0.02(-0.05)} & \makecell{ 0.54(0.69)\\ 0.29(0.26)\\ 12.1(12.2)\\ 0.70(0.84)\\ -0.02(0.00)} & \makecell{ 0.50(0.66)\\ 0.44(0.44)\\ 15.2(14.2)\\ 0.94(1.17)\\ 0.13(0.13)} & \makecell{ 0.54(0.76)\\ 0.27(0.28)\\ 10.5(10.7)\\ 1.02(1.20)\\ -0.01(-0.01)} & \makecell{ 0.43(0.67)\\ 0.18(0.12)\\ 10.8(10.6)\\ 0.83(0.90)\\ -0.13(-0.10)} & \makecell{ 0.57(0.73)\\ 0.21(0.15)\\ 10.5(10.6)\\ 0.78(1.03)\\ 0.02(-0.01)} & \makecell{ 0.62(0.77)\\ 0.49(0.52)\\ 13.0(13.1)\\ 0.95(0.97)\\ 0.02(0.03)} & \makecell{ 0.54(0.75)\\ 0.35(0.37)\\ 11.1(11.3)\\ 0.89(0.90)\\ -0.09(-0.07)} & \makecell{ 0.55(0.73)\\ 0.33(0.32)\\ 11.9(11.9)\\ 0.89(1.01)\\ -0.01(-0.00)} \\
\hline
        \end{tabular}
        \label{tab:quantitative-scene-wise-llff01}
    \end{table*}

    \begin{table*}
        \centering
        \caption{Per-scene performance of various models with three input views on LLFF dataset.
        The five rows show LPIPS, SSIM, PSNR, Depth MAE and Depth SROCC scores, respectively.
        The values within parenthesis show unmasked scores.}
        \begin{tabular}{l|c|c|c|c|c|c|c|c|c}
            \hline
            \textbf{\makecell{Model \textbackslash \\ Scene Name}} & \textbf{Fern} & \textbf{Flower} & \textbf{Fortress} & \textbf{Horns} & \textbf{Leaves} & \textbf{Orchids} & \textbf{Room} & \textbf{Trex} & \textbf{Average} \\
            \hline
            InfoNeRF     & \makecell{ 0.74(0.86)\\ 0.22(0.21)\\ 7.5(7.5)\\ 0.99(1.08)\\ 0.01(-0.07)} & \makecell{ 0.62(0.69)\\ 0.24(0.23)\\ 10.3(10.8)\\ 0.84(0.94)\\ -0.18(-0.11)} & \makecell{ 0.76(0.83)\\ 0.16(0.17)\\ 5.1(5.2)\\ 1.21(1.42)\\ 0.17(0.17)} & \makecell{ 0.63(0.76)\\ 0.18(0.18)\\ 8.9(8.9)\\ 1.07(1.26)\\ 0.04(0.05)} & \makecell{ 0.50(0.65)\\ 0.16(0.11)\\ 10.2(9.9)\\ 0.87(0.92)\\ -0.12(-0.10)} & \makecell{ 0.68(0.71)\\ 0.12(0.08)\\ 7.9(8.3)\\ 0.95(1.15)\\ 0.07(0.08)} & \makecell{ 0.75(0.83)\\ 0.29(0.29)\\ 8.7(8.9)\\ 1.04(1.07)\\ -0.11(-0.15)} & \makecell{ 0.69(0.79)\\ 0.19(0.18)\\ 8.6(8.7)\\ 0.93(0.96)\\ -0.03(-0.06)} & \makecell{ 0.67(0.77)\\ 0.20(0.19)\\ 8.4(8.5)\\ 1.00(1.11)\\ -0.01(-0.02)} \\
            \hline
            DietNeRF     & \makecell{ 0.68(0.80)\\ 0.33(0.32)\\ 11.9(12.0)\\ 0.86(0.95)\\ -0.06(-0.03)} & \makecell{ 0.58(0.67)\\ 0.33(0.31)\\ 13.2(13.1)\\ 0.73(0.84)\\ -0.16(-0.15)} & \makecell{ 0.63(0.70)\\ 0.40(0.41)\\ 12.6(12.4)\\ 1.00(1.22)\\ -0.08(-0.09)} & \makecell{ 0.60(0.75)\\ 0.28(0.29)\\ 10.9(11.1)\\ 1.03(1.21)\\ 0.03(0.06)} & \makecell{ 0.49(0.67)\\ 0.18(0.13)\\ 10.8(10.6)\\ 0.85(0.90)\\ -0.03(-0.03)} & \makecell{ 0.65(0.75)\\ 0.19(0.15)\\ 10.0(10.1)\\ 0.86(1.06)\\ -0.07(-0.08)} & \makecell{ 0.68(0.76)\\ 0.54(0.55)\\ 12.6(12.6)\\ 0.95(0.97)\\ 0.03(0.09)} & \makecell{ 0.60(0.72)\\ 0.36(0.35)\\ 11.8(11.9)\\ 0.88(0.91)\\ -0.16(-0.17)} & \makecell{ 0.61(0.73)\\ 0.34(0.33)\\ 11.8(11.8)\\ 0.91(1.02)\\ -0.06(-0.05)} \\
\hline
        \end{tabular}
        \label{tab:quantitative-scene-wise-llff02}
    \end{table*}

    \begin{table*}
        \centering
        \caption{Per-scene performance of various models with four input views on LLFF dataset.
        The five rows show LPIPS, SSIM, PSNR, Depth MAE and Depth SROCC scores, respectively.
        The values within parenthesis show unmasked scores.}
        \begin{tabular}{l|c|c|c|c|c|c|c|c|c}
            \hline
            \textbf{\makecell{Model \textbackslash \\ Scene Name}} & \textbf{Fern} & \textbf{Flower} & \textbf{Fortress} & \textbf{Horns} & \textbf{Leaves} & \textbf{Orchids} & \textbf{Room} & \textbf{Trex} & \textbf{Average} \\
            \hline
            InfoNeRF     & \makecell{ 0.71(0.80)\\ 0.20(0.19)\\ 9.6(9.8)\\ 0.95(1.03)\\ 0.02(0.00)} & \makecell{ 0.63(0.68)\\ 0.24(0.23)\\ 11.4(11.5)\\ 0.83(0.91)\\ -0.27(-0.21)} & \makecell{ 0.84(0.87)\\ 0.16(0.16)\\ 4.7(4.7)\\ 1.29(1.41)\\ 0.09(0.08)} & \makecell{ 0.68(0.77)\\ 0.19(0.19)\\ 8.9(8.8)\\ 1.18(1.27)\\ 0.08(0.09)} & \makecell{ 0.58(0.66)\\ 0.12(0.09)\\ 9.4(9.3)\\ 0.89(0.92)\\ -0.12(-0.11)} & \makecell{ 0.72(0.76)\\ 0.11(0.09)\\ 7.8(8.1)\\ 1.00(1.18)\\ 0.03(0.04)} & \makecell{ 0.72(0.78)\\ 0.45(0.45)\\ 11.9(11.9)\\ 1.05(1.06)\\ -0.06(-0.06)} & \makecell{ 0.69(0.80)\\ 0.26(0.26)\\ 9.9(10.1)\\ 0.97(0.97)\\ -0.13(-0.17)} & \makecell{ 0.70(0.77)\\ 0.23(0.22)\\ 9.2(9.2)\\ 1.04(1.11)\\ -0.04(-0.04)} \\
            \hline
            DietNeRF     & \makecell{ 0.66(0.80)\\ 0.36(0.35)\\ 12.7(12.9)\\ 0.85(0.93)\\ -0.11(-0.08)} & \makecell{ 0.62(0.69)\\ 0.31(0.29)\\ 12.6(12.6)\\ 0.76(0.84)\\ -0.05(-0.06)} & \makecell{ 0.67(0.70)\\ 0.42(0.42)\\ 12.8(12.6)\\ 1.07(1.19)\\ -0.04(-0.04)} & \makecell{ 0.66(0.78)\\ 0.30(0.30)\\ 10.9(10.8)\\ 1.13(1.22)\\ -0.00(0.02)} & \makecell{ 0.57(0.70)\\ 0.18(0.14)\\ 11.0(10.8)\\ 0.87(0.90)\\ -0.12(-0.10)} & \makecell{ 0.69(0.77)\\ 0.20(0.16)\\ 10.2(10.1)\\ 0.88(1.06)\\ -0.02(-0.02)} & \makecell{ 0.68(0.73)\\ 0.59(0.59)\\ 14.0(13.9)\\ 0.98(0.99)\\ 0.08(0.12)} & \makecell{ 0.64(0.75)\\ 0.34(0.35)\\ 10.9(11.2)\\ 0.91(0.92)\\ -0.10(-0.10)} & \makecell{ 0.65(0.74)\\ 0.35(0.34)\\ 11.9(11.8)\\ 0.95(1.03)\\ -0.04(-0.02)} \\
\hline
        \end{tabular}
        \label{tab:quantitative-scene-wise-llff03}
    \end{table*}

    \begin{table*}
        \centering
        \caption{Per-scene performance of ablated models with two input views on LLFF dataset.
        The five rows show LPIPS, SSIM, PSNR, Depth MAE and Depth SROCC scores, respectively.
        The values within parenthesis show unmasked scores.}
        \begin{tabular}{l|c|c|c|c|c|c|c|c|c}
            \hline
            \textbf{\makecell{Model \textbackslash \\ Scene Name}} & \textbf{Fern} & \textbf{Flower} & \textbf{Fortress} & \textbf{Horns} & \textbf{Leaves} & \textbf{Orchids} & \textbf{Room} & \textbf{Trex} & \textbf{Average} \\
            \hline
            SimpleNeRF   & \makecell{ 0.39(0.51)\\ 0.54(0.50)\\ 17.4(17.0)\\ 0.14(0.21)\\ \textbf{0.72}(0.68)} & \makecell{ 0.28(0.43)\\ \textbf{0.65}(0.53)\\ 20.1(16.9)\\ 0.10(0.25)\\ 0.94(0.72)} & \makecell{ 0.16(0.25)\\ 0.72(0.67)\\ \textbf{23.8}(\textbf{22.5})\\ 0.05(0.25)\\ \textbf{0.99}(\textbf{0.99})} & \makecell{ \textbf{0.26}(\textbf{0.42})\\ 0.64(0.54)\\ 19.6(17.1)\\ \textbf{0.18}(0.35)\\ \textbf{0.88}(\textbf{0.76})} & \makecell{ 0.25(0.44)\\ 0.45(0.30)\\ 15.2(13.5)\\ 0.31(0.44)\\ 0.64(0.38)} & \makecell{ \textbf{0.33}(0.41)\\ 0.52(0.37)\\ \textbf{16.3}(\textbf{14.7})\\ \textbf{0.15}(0.34)\\ \textbf{0.85}(\textbf{0.74})} & \makecell{ \textbf{0.27}(\textbf{0.35})\\ \textbf{0.83}(\textbf{0.77})\\ \textbf{22.5}(\textbf{19.5})\\ 0.12(\textbf{0.21})\\ 0.86(0.84)} & \makecell{ 0.28(0.39)\\ 0.68(0.58)\\ 18.3(16.8)\\ 0.13(0.19)\\ 0.78(0.73)} & \makecell{ \textbf{0.27}(\textbf{0.39})\\ \textbf{0.65}(\textbf{0.55})\\ \textbf{19.6}(\textbf{17.6})\\ \textbf{0.14}(0.28)\\ \textbf{0.85}(\textbf{0.75})} \\
\hline
        \end{tabular}
        \label{tab:quantitative-scene-wise-llff04}
    \end{table*}

    \begin{table*}
        \centering
        \caption{Per-scene performance of various models with two input views on RealEstate-10K dataset.
        The five rows show LPIPS, SSIM, PSNR, Depth MAE and Depth SROCC scores, respectively.
        The values within parenthesis show unmasked scores.}
        \begin{tabular}{l|c|c|c|c|c|c}
            \hline
            \textbf{Model \textbackslash \ Scene Name} & \textbf{0} & \textbf{1} & \textbf{3} & \textbf{4}  & \textbf{6}  & \textbf{Average} \\
            \hline
            InfoNeRF     & \makecell{ 0.5711(0.6035)\\ 0.4396(0.4356)\\ 11.70(11.68)\\ 2.2791(2.2552)\\ -0.0335(-0.0449)} & \makecell{ 0.4592(0.5016)\\ 0.6275(0.6339)\\ 16.20(15.61)\\ 2.0442(2.0017)\\ 0.1286(0.1617)} & \makecell{ 0.5959(0.6686)\\ 0.2549(0.2469)\\ 10.66(10.64)\\ 2.8117(2.9467)\\ 0.0727(0.0663)} & \makecell{ 0.7184(0.7605)\\ 0.3873(0.3936)\\ 10.70(10.83)\\ 2.3704(2.3495)\\ 0.0867(0.0712)} & \makecell{ 0.6175(0.6580)\\ 0.4616(0.4614)\\ 12.10(12.09)\\ 1.3912(1.6040)\\ 0.2014(0.2166)} & \makecell{ 0.5924(0.6384)\\ 0.4342(0.4343)\\ 12.27(12.17)\\ 2.1793(2.2314)\\ 0.0912(0.0942)} \\
            \hline
            DietNeRF     & \makecell{ 0.4738(0.5162)\\ 0.6335(0.6295)\\ 14.80(14.72)\\ 2.2511(2.2272)\\ 0.0047(0.0032)} & \makecell{ 0.3748(0.4149)\\ 0.8289(0.8387)\\ 21.04(20.58)\\ 1.9828(1.9476)\\ 0.5266(0.5465)} & \makecell{ 0.5748(0.6718)\\ 0.3356(0.3284)\\ 12.93(12.59)\\ 2.8672(3.0125)\\ 0.2821(0.3065)} & \makecell{ 0.4712(0.4986)\\ 0.6972(0.7016)\\ 18.47(18.50)\\ 2.0381(2.0190)\\ -0.1165(-0.0861)} & \makecell{ 0.2956(0.3296)\\ 0.7718(0.7617)\\ 23.05(22.74)\\ 0.9844(1.1973)\\ 0.4725(0.4559)} & \makecell{ 0.4381(0.4862)\\ 0.6534(0.6520)\\ 18.06(17.83)\\ 2.0247(2.0807)\\ 0.2339(0.2452)} \\
            \hline
            RegNeRF      & \makecell{ 0.3357(0.3480)\\ 0.6175(0.6010)\\ 16.77(16.51)\\ -- \\ -0.0420(-0.0269)} & \makecell{ 0.2898(0.3179)\\ 0.8226(0.8290)\\ 21.62(21.04)\\ -- \\ -0.3059(-0.2176)} & \makecell{ 0.4428(0.4944)\\ 0.3103(0.2973)\\ 14.16(13.88)\\ -- \\ 0.1796(0.1485)} & \makecell{ 0.4969(0.5390)\\ 0.6177(0.6148)\\ 17.46(17.13)\\ -- \\ 0.2285(0.1764)} & \makecell{ 0.4992(0.5421)\\ 0.5902(0.5901)\\ 15.68(15.79)\\ -- \\ 0.4986(0.4781)} & \makecell{ 0.4129(0.4483)\\ 0.5916(0.5864)\\ 17.14(16.87)\\ -- \\ 0.1118(0.1117)} \\
\hline
        \end{tabular}
        \label{tab:quantitative-scene-wise-realestate01}
    \end{table*}

    \begin{table*}
        \centering
        \caption{Per-scene performance of various models with three input views on RealEstate-10K dataset.
        The five rows show LPIPS, SSIM, PSNR, Depth MAE and Depth SROCC scores, respectively.
        The values within parenthesis show unmasked scores.}
        \begin{tabular}{l|c|c|c|c|c|c}
            \hline
            \textbf{Model \textbackslash \ Scene Name} & \textbf{0} & \textbf{1} & \textbf{3} & \textbf{4}  & \textbf{6}  & \textbf{Average} \\
            \hline
            InfoNeRF     & \makecell{ 0.6722(0.6983)\\ 0.3622(0.3593)\\ 9.74(9.73)\\ 2.3079(2.2937)\\ 0.0640(0.0560)} & \makecell{ 0.6151(0.6280)\\ 0.4916(0.4871)\\ 12.08(11.75)\\ 2.1419(2.1368)\\ 0.3196(0.3379)} & \makecell{ 0.6187(0.6621)\\ 0.1892(0.1848)\\ 9.08(9.23)\\ 2.8991(3.0636)\\ 0.0813(0.0884)} & \makecell{ 0.7524(0.7785)\\ 0.3749(0.3790)\\ 10.19(10.28)\\ 2.3476(2.3454)\\ 0.1210(0.1273)} & \makecell{ 0.6220(0.6562)\\ 0.4779(0.4797)\\ 11.74(11.87)\\ 1.4028(1.5757)\\ 0.3788(0.3875)} & \makecell{ 0.6561(0.6846)\\ 0.3792(0.3780)\\ 10.57(10.57)\\ 2.2198(2.2830)\\ 0.1929(0.1994)} \\
            \hline
            DietNeRF     & \makecell{ 0.4698(0.4893)\\ 0.6226(0.6166)\\ 16.44(16.36)\\ 2.1863(2.1710)\\ 0.0856(0.0728)} & \makecell{ 0.4979(0.5115)\\ 0.8046(0.8158)\\ 18.51(18.29)\\ 2.0866(2.0870)\\ -0.1136(-0.0664)} & \makecell{ 0.5759(0.6289)\\ 0.3235(0.3233)\\ 13.44(13.30)\\ 2.8594(3.0383)\\ 0.3096(0.3401)} & \makecell{ 0.4600(0.4719)\\ 0.7144(0.7151)\\ 18.98(18.95)\\ 1.9957(1.9918)\\ -0.2370(-0.2100)} & \makecell{ 0.3142(0.3413)\\ 0.7630(0.7514)\\ 22.65(22.54)\\ 1.0496(1.2234)\\ 0.0754(0.0823)} & \makecell{ 0.4636(0.4886)\\ 0.6456(0.6445)\\ 18.01(17.89)\\ 2.0355(2.1023)\\ 0.0240(0.0438)} \\
            \hline
            RegNeRF      & \makecell{ 0.3877(0.3952)\\ 0.6084(0.5950)\\ 16.22(15.99)\\ -- \\ -0.0854(-0.0849)} & \makecell{ 0.3179(0.3220)\\ 0.8178(0.8237)\\ 21.06(20.89)\\ -- \\ 0.2210(0.2376)} & \makecell{ 0.4885(0.5302)\\ 0.3008(0.2893)\\ 14.07(13.87)\\ -- \\ -0.0298(-0.0517)} & \makecell{ 0.5406(0.5635)\\ 0.6227(0.6221)\\ 17.61(17.60)\\ -- \\ 0.3273(0.2715)} & \makecell{ 0.3506(0.3699)\\ 0.7162(0.7087)\\ 20.35(20.28)\\ -- \\ -0.1459(-0.1349)} & \makecell{ 0.4171(0.4362)\\ 0.6132(0.6078)\\ 17.86(17.73)\\ -- \\ 0.0574(0.0475)} \\
\hline
        \end{tabular}
        \label{tab:quantitative-scene-wise-realestate02}
    \end{table*}

    \begin{table*}
        \centering
        \caption{Per-scene performance of various models with four input views on RealEstate-10K dataset.
        The five rows show LPIPS, SSIM, PSNR, Depth MAE and Depth SROCC scores, respectively.
        The values within parenthesis show unmasked scores.}
        \begin{tabular}{l|c|c|c|c|c|c}
            \hline
            \textbf{Model \textbackslash \ Scene Name} & \textbf{0} & \textbf{1} & \textbf{3} & \textbf{4}  & \textbf{6}  & \textbf{Average} \\
            \hline
            InfoNeRF     & \makecell{ 0.5948(0.6030)\\ 0.5023(0.5002)\\ 12.80(12.74)\\ 2.2648(2.2692)\\ 0.2109(0.2053)} & \makecell{ 0.7054(0.6927)\\ 0.3621(0.3621)\\ 8.76(8.60)\\ 2.0969(2.1482)\\ 0.2990(0.3040)} & \makecell{ 0.6889(0.6976)\\ 0.1767(0.1725)\\ 8.68(8.71)\\ 2.8542(3.0882)\\ 0.2874(0.2812)} & \makecell{ 0.7918(0.8068)\\ 0.3119(0.3176)\\ 8.76(8.79)\\ 2.3608(2.3658)\\ 0.1548(0.1696)} & \makecell{ 0.5447(0.5604)\\ 0.5683(0.5625)\\ 14.12(14.12)\\ 1.3605(1.4993)\\ 0.3227(0.3369)} & \makecell{ 0.6651(0.6721)\\ 0.3843(0.3830)\\ 10.62(10.59)\\ 2.1874(2.2742)\\ 0.2549(0.2594)} \\
            \hline
            DietNeRF     & \makecell{ 0.5095(0.5189)\\ 0.6163(0.6125)\\ 15.78(15.63)\\ 2.2313(2.2346)\\ 0.0064(0.0055)} & \makecell{ 0.4664(0.4628)\\ 0.8352(0.8399)\\ 19.91(19.64)\\ 2.0104(2.0647)\\ 0.1436(0.1283)} & \makecell{ 0.6182(0.6399)\\ 0.3243(0.3198)\\ 13.29(13.27)\\ 2.7825(3.0206)\\ 0.3925(0.3897)} & \makecell{ 0.4881(0.4966)\\ 0.7157(0.7156)\\ 19.06(19.01)\\ 1.9930(1.9969)\\ -0.1838(-0.1637)} & \makecell{ 0.3442(0.3589)\\ 0.7598(0.7495)\\ 21.99(21.93)\\ 1.1820(1.3199)\\ 0.1363(0.1458)} & \makecell{ 0.4853(0.4954)\\ 0.6503(0.6475)\\ 18.01(17.89)\\ 2.0398(2.1273)\\ 0.0990(0.1011)} \\
            \hline
            RegNeRF      & \makecell{ 0.4215(0.4252)\\ 0.5922(0.5824)\\ 16.29(16.09)\\ -- \\ -0.0032(-0.0018)} & \makecell{ 0.3498(0.3461)\\ 0.8249(0.8256)\\ 21.14(20.98)\\ -- \\ 0.2630(0.2456)} & \makecell{ 0.5737(0.5903)\\ 0.3036(0.2936)\\ 13.93(13.91)\\ -- \\ 0.2565(0.2652)} & \makecell{ 0.5450(0.5581)\\ 0.6484(0.6479)\\ 18.49(18.48)\\ -- \\ 0.4133(0.4104)} & \makecell{ 0.2678(0.2719)\\ 0.7595(0.7497)\\ 21.86(21.78)\\ -- \\ -0.2187(-0.2214)} & \makecell{ 0.4316(0.4383)\\ 0.6257(0.6198)\\ 18.34(18.25)\\ -- \\ 0.1422(0.1396)} \\
\hline
        \end{tabular}
        \label{tab:quantitative-scene-wise-realestate03}
    \end{table*}

    \begin{table*}
        \centering
        \caption{Per-scene performance of ablated models with two input views on RealEstate-10K dataset.
        The five rows show LPIPS, SSIM, PSNR, Depth MAE and Depth SROCC scores, respectively.
        The values within parenthesis show unmasked scores.}
        \begin{tabular}{l|c|c|c|c|c|c}
            \hline
            \textbf{Model \textbackslash \ Scene Name} & \textbf{0} & \textbf{1} & \textbf{3} & \textbf{4}  & \textbf{6}  & \textbf{Average} \\
            \hline
            SimpleNeRF   & \makecell{ \textbf{0.0276}(\textbf{0.0369})\\ \textbf{0.9662}(\textbf{0.9532})\\ 34.94(31.89)\\ \textbf{0.3299}(\textbf{0.3764})\\ \textbf{0.8490}(\textbf{0.8007})} & \makecell{ \textbf{0.0273}(\textbf{0.0392})\\ \textbf{0.9823}(\textbf{0.9728})\\ \textbf{38.36}(\textbf{33.80})\\ \textbf{0.2283}(\textbf{0.3202})\\ \textbf{0.9733}(\textbf{0.9309})} & \makecell{ 0.1913(0.2130)\\ 0.5968(0.5590)\\ 19.23(18.65)\\ 0.7286(\textbf{1.0194})\\ 0.9414(0.8582)} & \makecell{ 0.0300(0.0336)\\ 0.9592(0.9511)\\ 38.51(34.93)\\ \textbf{0.0730}(\textbf{0.0960})\\ \textbf{0.9624}(\textbf{0.9363})} & \makecell{ \textbf{0.0413}(\textbf{0.0500})\\ \textbf{0.9663}(\textbf{0.9553})\\ 34.47(32.24)\\ \textbf{0.2747}(\textbf{0.4803})\\ \textbf{0.8812}(\textbf{0.8647})} & \makecell{ \textbf{0.0635}(\textbf{0.0745})\\ \textbf{0.8942}(\textbf{0.8783})\\ \textbf{33.10}(30.30)\\ \textbf{0.3269}(\textbf{0.4584})\\ \textbf{0.9215}(\textbf{0.8781})} \\
            \hline
            \makecell[l]{SimpleNeRF w/o \\ Points Augmentation} & \makecell{ 0.0334(0.0426)\\ 0.9618(0.9487)\\ 34.26(31.31)\\ 0.3632(0.4080)\\ 0.8279(0.7849)} & \makecell{ 0.0387(0.0543)\\ 0.9768(0.9672)\\ 37.15(32.80)\\ 0.3145(0.3992)\\ 0.9211(0.8817)} & \makecell{ 0.1988(0.2215)\\ 0.5927(0.5540)\\ 19.28(18.76)\\ 0.7602(1.0816)\\ 0.9074(0.7931)} & \makecell{ 0.0315(0.0365)\\ 0.9582(0.9492)\\ 38.27(34.50)\\ 0.0862(0.1146)\\ 0.9601(0.9296)} & \makecell{ 0.0736(0.0898)\\ 0.9533(0.9418)\\ 33.75(31.87)\\ 0.3733(0.5510)\\ 0.8699(0.8540)} & \makecell{ 0.0752(0.0889)\\ 0.8886(0.8722)\\ 32.54(29.85)\\ 0.3795(0.5109)\\ 0.8973(0.8487)} \\
\hline
        \end{tabular}
        \label{tab:quantitative-scene-wise-realestate04}
    \end{table*}

\end{document}